\pdfoutput=1

\documentclass[preprint,12pt]{article}

\usepackage[accsupp]{axessibility}

\usepackage{amsmath}
\usepackage{amsthm}
\usepackage{makeidx}
\usepackage{latexsym}
\usepackage{amssymb}

\usepackage{slashbox}

\usepackage[caption=false,font=footnotesize]{subfig}
\usepackage{floatflt}
\usepackage{graphicx}
\usepackage{bm}
\usepackage{epsfig}
\usepackage{epsf}

\usepackage{hyperref}
\usepackage{float}
\usepackage[ruled,vlined,linesnumbered,inoutnumbered]{algorithm2e}

\newtheorem{remark}{Remark}

\theoremstyle{remark}

\usepackage{geometry}
\date{}

\title{A Bayesian approach for initialization of weights in backpropagation neural net with application to character recognition}
\author{Nadir Murru\footnote{Corresponding author: nadir.murru@gmail.com}, Rosaria Rossini\\
Department of Mathematics, University of Turin\\ Via Carlo Alberto 8/10, Turin, Italy\\ nadir.murru@unito.it, rossini@di.unito.it}

\begin{document}

\maketitle

\begin{abstract}
Convergence rate of training algorithms for neural networks is heavily affected by initialization of weights. 
In this paper, an original algorithm for initialization of weights in backpropagation neural net is presented with application to character recognition. The initialization method is mainly based on a customization of the Kalman filter, translating it into Bayesian statistics terms. A metrological approach is used in this context considering weights as measurements modeled by mutually dependent normal random variables. The algorithm performance is demonstrated by reporting and discussing results of simulation trials. Results are compared with random weights initialization and other methods. The proposed method shows an improved convergence rate for the backpropagation training algorithm.
\end{abstract}

\textbf{keywords:}
backpropagation algorithm; Bayesian statistics; character recognition; Kalman filter; neural network.

\section{Introduction}

In the last decades, neural networks have generated much interest 
both from a theoretical point of view and for their several applications in complex problems, 
such as function approximations, data processing, robotics, computer numerical control. Moreover, neural nets are particularly exploited in pattern recognition and consequently can be conveniently used 
in the realization of Optical Character Recognition (OCR) software. 

An artificial neural network (ANN) is a mathematical model designed as the structure of the nervous system.   
The model was presented for the first time by McCulloch and Pitts \cite{MCP} and
involves four main components: a set of nodes (neurons), their connections (synapses), an activation function that determines the output of each node and a set of weights associated to the connections.

Initialization of weights heavily affects performances of feedforward neural networks \cite{Thimm},
as a consequence many different initialization methods have been studied. Since neural nets are applied to many different complex problems, these methods have fluctuating performances. For this reason, random weight initialization is still the most used method also due to its simplicity. 
Thus, the study of new weight initialization methods is an important research field in order to improve application of neural nets and deepen their knowledge.

In this paper we focus on feedforward neural nets trained by using the Backpropagation (BP) algorithm, which is a widely used method of training. It is well--known that convergence of BP neural net is heavily affected by initial weights \cite{Bat}, \cite{Thimm}, \cite{Liu2}, \cite{Adam2}. 

Different initialization techniques have been proposed for feedforward neural nets, such as adaptive step size method \cite{Sch} and partial least squares method \cite{Liu}. 
Hsiao et al. \cite{Hsiao} applied the partial least squares method to BP network. 
Duch et al. \cite{Duch} investigated the optimal initialization of multilayered perceptrons by means of clusterization techniques. Varnava and Meade \cite{Var} constructed an initialization method for feedforward neural nets by using polynomial bases.

Kathirvalavakumar and Subavathi \cite{Kat} proposed a method that improves convergence rate exploiting Cauchy inequality and performing a sensitivity analysis. An interval based weight initialization method is presented in \cite{Sodhi}, where authors used the resilient BP algorithm for testing. Adam et al. \cite{Adam} treated the problem of initial weights in terms of solving a linear interval tolerance problem and tested their method on neural networks trained with BP algorithm.

Yam et al. \cite{Yam} evaluated optimal initial weights by using a least squares method that minimizes the initial error allowing convergence of neural net by a reduced number of steps. The method is tested on BP neural net with application to character recognition. Other different approaches can be found in \cite{Erdo}, \cite{Asadi}, \cite{Kus}, \cite{Pet} where authors focused on BP artificial neural network. 

A comparison among several weight initialization methods can be found in \cite{Red}, where the authors tested methods on BP network with hyperbolic tangent transfer function.

In this paper, we propose a novel approach based on a Bayesian estimation of initial weights. 
Bayesian estimation techniques are widely used in many different contexts. For instance, in \cite{Der1} authors developed a customization of the Kalman filter, translating it into Bayesian statistics terms. The purpose of this customization was to address metrological problems. Here, we extend such an approach in order to evaluate an optimized set of initial weights for BP neural net with sigmoidal transfer function. Through several simulations we show the effectiveness of our approach in the field of character recognition.

The paper is structured as follows. In Section \ref{sec:bp}, we briefly recall the BP training algorithm. 
In Section \ref{sec:bayes} we discuss a novel approach for weight initialization in BP neural nets using a Bayesian approach derived by a customization of the Kalman filter. 
In Section \ref{sec:exp}, we discuss the setting of some parameters and we show experimental results on the convergence of BP neural net in character recognition. Our Bayesian weight initialization method is compared with classical random initialization and other methods. 
A sensitivity analysis on some parameters is also presented here.  
Section \ref{sec:conc} concludes the paper.

\section{Overview of Backpropagation training algorithm} \label{sec:bp}
In this section we present an overview of the BP training algorithm introducing some notation. 

Let us consider a feedforward neural network with \(L\) layers. Let \(N(i)\) be the number of neurons in layer \(i\), for \(i=1,...,L\), and \(w^{(k)}\) be the weight matrix \(N(k)\times N(k-1)\) corresponding to connections among neurons in layers \(k\) and \(k-1\), for \(k=2,...,L\). In other words, \(w^{(k)}_{ij}\) is the weight of connection between \(i\)--th neuron in layer \(k\) and \(j\)--th neuron in layer \(k-1\). In the following, we will consider biases equal to zero for seek of simplicity.

Artificial neural networks are trained over a set of inputs so that the neural net provides a fixed output for a given training input. Let us denote \(X\) the set of training inputs and \(n=\lvert X \rvert\) the number of different training inputs. An element \(\textbf x\in X\) is a vector (e.g., a string of bit 0 and 1) whose length is usually equals to \(N(1)\). In the following, bold symbols will denote vectorial quantities.

Let \(a_i^{(k,\textbf x)}\) be the activation of neuron \(i\) in layer \(k\) given the input \(\textbf{x}\):
\[\begin{cases} a_i^{(1,\textbf x)}=\sigma(x_i) \cr  a_i^{(k,\textbf x)}=\sigma\left(\sum_{j=1}^{N(k-1)}w_{ij}^{(k)}a_j^{(k-1,\textbf{x})}\right),\quad k=2,...,L \end{cases},\]
where $\sigma$ is the transfer function. In the following, $\sigma$ will be the sigmoidal function. Moreover, let us denote $z_i^{(k,\textbf x)}$ the weighted input to the activation function for neuron $i$ in layer $k$, given the input $\textbf x$:
$$\begin{cases} z_i^{(1,\textbf x)}=x_i \cr  z_i^{(k,\textbf x)}=\sum_{j=1}^{N(k-1)}w_{ij}^{(k)}a_j^{(k-1,\textbf{x})},\quad k=2,...,L \end{cases}.$$
Using vectorial notation, we have
$$\begin{cases} \textbf{z}^{(1,\textbf x)}=\textbf{x}, \quad \textbf{a}^{(1,\textbf{x})}=\sigma(\textbf z) \cr \textbf{z}^{(k,\textbf x)}=w^{(k)}\textbf{a}^{(k-1,\textbf x)},\quad \textbf{a}^{(k,\textbf x)}=\sigma(\textbf{z}^{(k,\textbf x)}),\quad k=2,...,L  \end{cases}.$$
Finally, let $\textbf y^{(\textbf x)}$ be the desired output of the neural network corresponding to input $\textbf x$. In other words, we would like that $\textbf a^{(L,\textbf x)}=\textbf y^{(\textbf x)}$, when neural net processes input $\textbf x$. Clearly, this depends on weights $w^{(k)}_{ij}$ and it is not possible to know their correct values a priori. Thus, it is usual to randomly initialize values of weights and use a training algorithm in order to adjust their values. In Algorithm \ref{alg:bp}, the BP training algorithm is described.
\begin{algorithm}[!htb]
\DontPrintSemicolon
\KwData{\\
				$L$ number of layers\\
				$N(k)$ number of neurons in layer $k$, for $k=1,...,L$\\
				$w_{ij}^{(k)}$ initial weights, for $i=1,...,N(k)$, $j=1,...,N(k-1)$, $k=2,...,L$\\
				$X$ set of training inputs, $n=\lvert X \rvert$\\
				$\textbf y^{(\textbf x)}$ desired output for all training inputs $\textbf x\in X$\\
				$\eta$ learning rate}
\KwResult{$w_{ij}^{(k)}$ final weights, for $i=1,...,N(k)$, $j=1,...,N(k-1)$, $k=2,...,L$, such that $\textbf a^{(L,\textbf x)}=\textbf y^{(\textbf x)}$, $\forall \textbf x\in X$}
\Begin{
\While{$\exists \textbf{x}\in X: \textbf a^{(L,\textbf x)}\not=\textbf y^{(\textbf x)}$}{
\For(\tcp*[f]{for each training input}){$\textbf x\in X$}{
		$\textbf a^{(1,\textbf x)}=\sigma(\textbf x)$ \\
		\For{k=2,...,L}{
				$\textbf z^{(k,\textbf x)}=w^{(k)}\textbf a^{(k-1,\textbf x)},\quad \textbf a^{(k,\textbf x)}=\sigma(\textbf z^{(k,\textbf x)})$
		}
		$\textbf d^{(L,\textbf x)}=(\textbf a^{(L,\textbf x)}-\textbf y^{(\textbf x)})\odot \sigma'(\textbf z^{(L,\textbf x)})$, \tcp*{$\odot$ componentwise product}
		\For{k=L-1,...,2}{
				$\textbf d^{(k,\textbf x)}=((w^{(k+1)})^T\textbf d^{(k+1,\textbf x)})\odot\sigma'(\textbf z^{(k,\textbf x)})$ \tcp*{right superscript $T$ stands for transpose operator}
				}
}
\For{k=L,...,2}{$w^{(k)}=w^{(k)}-\frac{\eta}{n}\sum_{\textbf x \in X} \textbf d^{(k,\textbf x)}(\textbf a^{(k-1,\textbf x)})^T$
}}
}
\caption{Backpropagation training algorithm \label{alg:bp}}
\end{algorithm}

\section{Bayesian weight initialization based on a customized Kalman filter technique} \label{sec:bayes}
The Kalman filter \cite{Kalman} is a well--established method to estimate the state $\textbf w_t$ of a dynamic process at each time $t$. 
The estimation $\mathbf{\tilde w}_t$ is obtained  balancing prior estimations and measurements of the process $\textbf w_t$ by means of the Kalman gain matrix. This matrix is constructed in order to minimize the mean--square--error $\mathbb E[(\mathbf{\tilde w}_t-\textbf w_t)(\mathbf{\tilde w}_t-\textbf w_t)^T]$. Estimates attained by Kalman filter are optimal under such diverse criteria, like least-squares or minimum-mean-square-error, and its practice is developed with application to several fields.
 
The Kalman filter has been successfully used with neural networks \cite{Book}. In this context, training of neural networks is treated as a non--linear estimating problem and consequently the extended Kalman filter is usually exploited in order to derive new training algorithms. Many modifications of the extended Kalman filter exist, thus different algorithms have been developed as, e.g., in \cite{Sin}, \cite{Wat}, \cite{Hei}, \cite{Riv}. However, extended Kalman filter is computationally complex and needs tuning several parameters that makes its implementation a difficult problem (see, e.g., \cite{Julier}). 

In this section, we show that classical Kalman filter could be used in place of the extended version, constructing a simplified Kalman filter used in combination with BP algorithm in order to reduce computational costs. The motivations about using Kalman filter and proposing a novel approach can be summarized as follows: Kalman filter is widespread in several applied fields in order to optimize performances (including neural networks); it produces optimal estimations under diverse and well--established criteria; it has been used with neural networks mainly in the extended version with the problems above specified.
 
Let the dynamic of the process be described by the following equation:
\begin{equation} \label{eq:process} \textbf w_{t+1}=A_t\textbf w_t+B_t\textbf u_t+\textbf p_t \end{equation} 
where $\textbf u_t, \textbf p_t$ are the optional control input and the white noise, respectively. Matrices $A_t, B_t$ are used
to relate the process state at the step $t+1$ to the $t$--th process state and to the $t$--th control input, respectively. 

We now introduce the (direct) measurement values of the process $\textbf m_t$ as:
$$\textbf m_t=\textbf w_t+\textbf r_t$$
where $\textbf r_t$ represents measurements uncertainty. 
Given that, a simplified version of the estimation $\mathbf{\tilde w}_t$  produced by the Kalman filter can be represented as follows:
\begin{equation} \label{eq:kf} \mathbf{\tilde w}_t=\textbf w^-_t+K_t(\textbf m_t-\textbf w^-_t) \end{equation}
where $K_t$ is the Kalman gain matrix and
$$\textbf w_t^-=A_{t-1}\mathbf{\tilde w}_{t-1}+B_{t-1}\textbf u_{t-1}$$
for a given initial prior estimation $\textbf w_0^-$ of $\textbf w_0$.

As stated in the introduction, the Kalman filter has been applied to dimensional metrology by D'Errico and Murru in \cite{Der1}. 
The aim of the authors was to minimize the error of measurement instrumentations deriving a simplified version of the Kalman gain matrix by using the Bayes theorem and considering components of each state of the process $\textbf w_t$ as mutually independent normal random variables. 

In this section, we extend such an approach in order to optimize weights initialization of neural networks. 
In particular, we introduce a possible correlations among components of $\textbf w_t$ and we consider the weights as processes whose measurements are provided by random sampling. Furthermore, in the following section, we will specify the construction of some covariance matrices necessary to apply the Kalman filter in this context.

Using the above notation, let $\textbf W_t$ and $\textbf M_t$ be multivariate random variables such that 
\begin{equation} \label{eq:model} f(\textbf W_t)=\mathcal N(\textbf w_t^-,Q_t),\quad f(\textbf M_t|\textbf W_t)=\mathcal N(\textbf m_t, R_t),\quad 0\leq t \leq t_{max} \end{equation}
where $\mathcal N(\mathbf{\mu},\Sigma)$ is a Gaussian multivariate probability density function with mean $\mathbf{\mu}$ and covariance matrix $\Sigma$. In \eqref{eq:model}, the random variable $\textbf W_t$ models prior estimations and $Q_t$ is the covariance matrix whose diagonal entries represent their uncertainties and non--diagonal entries are correlations between components of $\textbf w_t^-$. Similarly, $\textbf M_t|\textbf W_t$ models measurements and $R_t$ is the covariance matrix whose entries describe same information of $Q_t$ related to $\textbf m_t$.

The Bayes theorem states that
$$f(\textbf W_t|\textbf M_t)=\cfrac{f(\textbf M_t|\textbf W_t)f(\textbf W_t)}{\int_{-\infty}^{+\infty}f(\textbf M_t|\textbf W_t)f(\textbf W_t) d\textbf W_t}$$
where $f(\textbf W_t|\textbf M_t)$ is called the posterior density, $f(\textbf W_t)$ the prior density and $f(\textbf M_t|\textbf W_t)$ the likelihood.
We have
$$f(\textbf W_t|\textbf M_t)\propto \mathcal N(\textbf w_t^-,Q_t)\mathcal N(\textbf m_t, R_t)=\mathcal N(\mathbf{\tilde w}_t,P_t)$$
where
$$ \mathbf{\tilde w}_t=(Q_t^{-1}+R_t^{-1})^{-1}(Q_t^{-1}\textbf w_t^-+R_t^{-1}\textbf m_t), \quad P_t=(Q_t^{-1}+R_t^{-1})^{-1}. $$
In metrological terms, diagonal entries of $P_t$ can be used for type B uncertainty treatment (see the guide \cite{guide}) and the expected value of the posterior Gaussian $f(\textbf W_{t_{max}}|\textbf M_{t_{max}})$ is the final estimate of the process.

We can apply this technique to weights initialization considering processes $\textbf w_t(k)$, for $k=2,...,L$, as non--time--varying quantities, i.e.,
\begin{equation} \label{eq:process-weight} \textbf w_{t+1}(k)=\textbf w_t(k)+\textbf p_t(k) \end{equation}
whose components are the unknown values of weights $w^{(k)}$, for $k=2,...,L$, of the neural net such that $\textbf a^{(L,\textbf x)}=\textbf y^{(\textbf x)}$. Eq. \eqref{eq:process-weight} is the simplified version of \eqref{eq:process}, i.e., describes the dynamics of our processes.

The goal is to provide an estimation of initial weights to reduce the number of steps that allows convergence of BP neural net. 

Thus, for each set $w^{(k)}$ we consider initial weights as unknown processes and we optimize randomly generated weights (which we consider as measurements of the processes) with the above approach. 
In these terms, we derive an optimal initialization of weights by means of the following equations:
\begin{equation} \label{eq:kfc} \begin{cases}\mathbf{\tilde w}_t=(Q_t^{-1}+R_t^{-1})^{-1}(Q_t^{-1}\textbf w_t^-+R_t^{-1}\textbf m_t) \cr Q_{t+1}=(Q_t^{-1}+R_t^{-1})^{-1} \cr \textbf w_{t+1}^-=\mathbf{\tilde w}_{t} \end{cases} \end{equation}
for $t$ varying from 0 to $t_{max}$ and for each set of weights $w^{(k)}$. For the sake of simplicity we omitted dependence from $k$ in the above equations. In Equations \eqref{eq:kfc}, the initial state $\textbf w^-_0$ of $\textbf w_t$ is a prior estimation of $\textbf w_0$ that should be provided. Moreover, covariance matrices $Q_0$ and $R_t$ must be set in a convenient way. First equation in \eqref{eq:kfc} is the metrological realization of the Kalman--based equation \eqref{eq:kf}. From previous equations, we derive the Kalman gain matrix as
$$K_t=(Q_t^{-1}+R_t^{-1})^{-1}R_t^{-1}.$$
Indeed, we have that $I-K_t=(Q_t^{-1}+R_t^{-1})^{-1}Q_t^{-1}$, where $I$ is the identity matrix. 

In the following section we discuss the setting of these parameters and 
we also provide the results about the comparison of our approach to random initialization with application to character recognition.

\section{Numerical results} \label{sec:exp}

In this section, we explain the process of our weights initialization and the involved parameters with particular attention to the structure of the  covariance matrices, Section \ref{sec:param}. To evaluate performances of the BP algorithm with random weights initialization (RI) against Bayesian weights initialization (BI) provided by Algorithm \ref{alg:winit}, we apply neural nets in character recognition. In particular, we discuss the results of our experimental evaluation about the comparison of our approach with a random approach initialization conduct in a field of printed character recognition, taking into account convergence rate, Section \ref{sec:printed}. In this section we use a neural net with 3 layers and sigmoidal activation function. Afterwards, we train BP neural nets (with 3 and 5 layers, using both sigmoidal and hyperbolic tangent activation functions) on the MNIST database for the recognition of handwritten digits, Section \ref{sec:mnist}. In these simulations, we also take into account classification accuracy. Finally, we compare BI method with other methods in Section \ref{sec:comparison}.
These experiments show the advantage that our approach provides in terms of number of steps used to train the artificial neural network.

\subsection{Parameters of weights initialization algorithm}\label{sec:param}
The method of weights initialization described in Section \ref{sec:bayes} is presented in Algorithm \ref{alg:winit}.

\begin{algorithm}[ht]
\DontPrintSemicolon
\KwData{\\
				$L$ number of layers\\
				$N(k)$ number of neurons in layer $k$, for $k=1,...,L$\\
				$X$ set of training inputs, $n=\lvert X \rvert$\\
				$\textbf y^{(\textbf x)}$ desired output for all training inputs $\textbf x\in X$\\
				$Q_0(k)$, for $k=2,...,L$\\
				$\textbf w^{-}_0(k)$ prior estimation of $\mathbf{\bar{w}}^{(k)}$, for $k=2,...,L$\\
				$\textbf m_0(k)$ measurement of $\mathbf{\bar{w}}^{(k)}$, for $k=2,...,L$ \\ 
				$R_0(k)$, for $k=2,...,L$
				}
\KwResult{$\mathbf{\tilde w}_2(k)$, optimized initial weights for backpropagation algorithm, for $k=2,...,L$}
\Begin{
\For(\tcp*[f]{for each set of weights}){k=2,...,L}{
\For(){$t=0,1,2$}{$\mathbf{\tilde w}_t(k)=(Q_t^{-1}(k)+R_t^{-1}(k))^{-1}(k)(Q_t^{-1}(k)\textbf w_t^-(k)+R_t^{-1}(k)\textbf m_t(k))$\\
$Q_{t+1}(k)=(Q_t^{-1}(k)+R_t^{-1}(k))^{-1}$\\
$w_{t+1}^-(k)=\mathbf{\tilde w}_{t}(k)$\\
$\textbf m_{t+1}(k)=Rnd(-h,h)$ \tcp*{$Rnd(-h,h)$ random sampling in the interval $(-h,h)$}
$(R_{t+1}(k))_{ii}=\cfrac{1}{N(k)N(k-1)}\sum_{\textbf x\in X}\| \textbf{d}^{(k,\textbf x)} \|^2$, $\forall i$\\
$(R_{t+1}(k))_{lm}=0.7$, $\forall l,m$ 
}}
}
\caption{Weights initialization algorithm based on Kalman filter \label{alg:winit}}
\end{algorithm}

Since we do not have any prior knowledge about processes $\mathbf{w}(k)$, the random variable $\textbf W_0(k)$, which models initial prior estimation, is initialized with the normal distribution $\mathcal N(\textbf 0,\cfrac{1}{\epsilon}I)$, where $\epsilon$ is a small quantity. In our  simulations, we will use a fixed $\epsilon=10^{-5}$. Note that such an initialization is a standard \cite{Sin}. 

Measurements $\textbf m_t(k)$ are obtained by randomly sampling in the real interval $(-h,h)$, for all $t$.  Usually the value of $h$  depends on the specific problem where neural net is applied. Then, we provide a sensitivity analysis on this parameter in the discussion of the results.

The covariance matrix $R_t(k)$ is a symmetric matrix whose entries outside the main diagonal are set equal to 0.7. This choice is based on a sensitivity analysis involving the Pearson coefficient (about correlations of weights) that improves performance of our algorithm. In \cite{Der1}, diagonal entries of covariance matrices were used to describe uncertainty of measurements. In our context, high values of $(R_t(k))_{ii}$ reflect bad accuracy of $(\textbf{m}_t(k))_i$, i. e., this weight affects output of the neural net being far from the desired output. Thus, we can use values of $\textbf{d}^{(k,\textbf x)}$ to measure inaccuracy of $\textbf{m}_t(k)$ as follows:
$$(R_t(k))_{ii}=\cfrac{1}{N(k)N(k-1)}\sum_{\textbf x\in X}\| \textbf{d}^{(k,\textbf x)} \|^2,\quad \forall i, \forall k,$$
where $\|\cdot\|$ stands for the Euclidean norm. Quantity $\| \textbf{d}^{(k,\textbf x)} \|^2$ expresses distance from output and desired output of $k$--th layer, given the input $\textbf x$. The sum over all $\textbf x\in X$ measures the total inaccuracy of the output of $k$--th layer. We divide by the number of weights connecting neurons in layers $k-1$ and $k$ so that $(R_t(k))_{ii}$ represents in mean the inaccuracy of a single weight connecting a neuron in layer $k-1$ with a neuron in layer $k$. 

Finally, we iterate Eqs. \eqref{eq:kfc} for a small number of times. Indeed, entries of $Q_t$ rapidly decrease with respect to $R_t$ by means of second equation in \eqref{eq:kfc}. Consequently after a few steps, in first equation of \eqref{eq:kfc}, $\textbf{w}^-_t$ has much greater weight than $\textbf m_t$ so that improvements of $\mathbf{\tilde w}_t$ could not be significative. In our simulations, we fixed a threshold of  $t_{max}=2$ in order to reduce number of iterations of our algorithm (and consequently number of operations) but obtaining a significant reduction of the step number in the BP algorithm.

\begin{remark}
The computational complexity to implement the classical Kalman filter is polynomial (see, e.g., \cite{Fault} p. 226). Our customization described in Algorithm \ref{alg:winit} is faster for the following reasons:
\begin{itemize}
\item it involves a less number of operations (matrix multiplications) than usual Kalman filter;
\item in the Kalman filter the most time consuming operation is given by the evaluation of inverse of matrices. In our case, this can be performed in a fast way, since we deal with circulant matrices, i.e., matrices where each row is a cyclic shift of the row above it. It is well--known that inverse of circulant matrices can be evaluated in a very fast way. Indeed, they can be diagonalized by using the Discrete Fourier Transform (\cite{Toe}, p. 32); the Discrete Fourier Transform and the inverse of a diagonal matrix are immediate to evaluate.
\end{itemize}
Thus, our algorithm is faster than classical Kalman filter, moreover it is iterated for a low number of steps ($t_{max}$=2). Surely, Algorithm \ref{alg:winit} has a time complexity greater than random initialization. However, looking at BP Algorithm \ref{alg:bp}, we can observe that Algorithm \ref{alg:winit} involves similar operations (i.e., matrix multiplications or multiplications between matrices and vectors) in a minor quantity as well as it needs a smaller number of cycles. Furthermore, in the following sections, we will see that weights initialization by means of Algorithm \ref{alg:winit} generally leads to a noticeable decrease of steps necessary for the convergence of the BP algorithm with respect to random initialization. Thus, using Algorithm \ref{alg:winit} we can reach a faster convergence, in terms of time, of the BP algorithm than using random initialization.
\end{remark}

\subsection{Experiments on latin printed characters}\label{sec:printed}

In this section we train the neural network in order to recognize latin printed characters using both BI and RI methods and we compare these results.

\begin{figure}[ht] 
\centering
\includegraphics[scale=0.25]{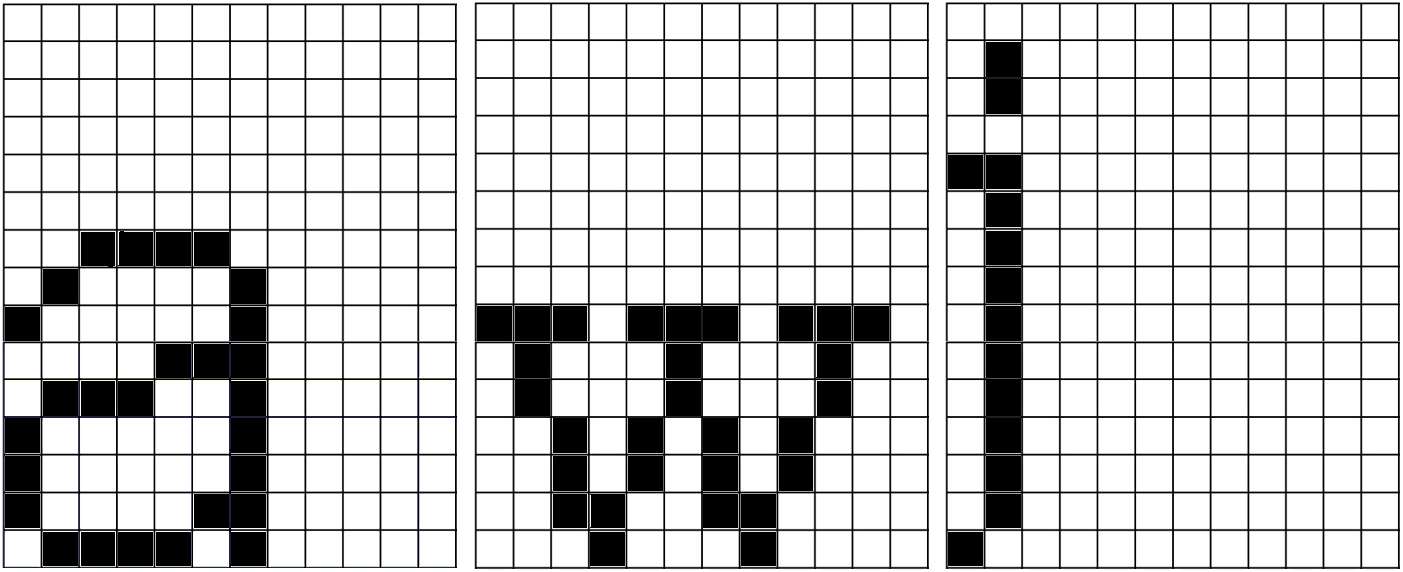}
\caption{Example of characters of the dataset: letter ''a'', font arial, pt 12; letter ''w'', font times new roman, pt 12; letter ''j'', font georgia, pt 12}
\label{fig:lettera}
\end{figure}

The set $X$ of training inputs is composed by 26 characters of the alphabet for 5 different fonts (Arial, Courier, Georgia, Times New Roman, Verdana) with 12 pt. Thus, we have $n=130$ different inputs. The characters are considered as binary images contained in $15\times12$ rectangles. Thus, an element $\textbf x\in X$ is a vector of length $15\cdot12=180$ with components 0 or 1. Figure \ref{fig:lettera} shows an example of characters of our dataset. A white pixel is coded with 0, a black pixel is coded with 1. The corresponding vector is constructed reading the matrix row--by--row (from left to right, from down to top).

For the experiment presented here, we use a neural net with $L=3$ layers, $N(1)=15\cdot12=180$, $N(3)=26$. 
Conventionally, size of first layer is equal to size of training inputs and size of last layer is equal to the number of different desired outputs. In our case, last layer has 26 neurons, as the characters of the latin alphabet. The desired output $\textbf y^{(\textbf x)}$ is the vector $(1,0,0,...,0)$, of length 26, when input $\textbf x$ is the character \emph{a} (for any font), is the vector $(0,1,0,...,0)$ when the input is the character \emph{b}, etc.

For comparison purposes, simulations are performed for different values of parameters $N(2)$, $h$, and $\eta$. We recall that $N(2)$ is the number of neurons in layer 2, $(-h,h)$ is the interval where weights are sampled, and $\eta$ is the learning rate. To the best of our knowledge these parameters have not a standard initialization, see, e.g., \cite{Rojas}.  

For each  combination of $N(2), h, \eta$, we train the neural net with RI for 1000 different times and we evaluate the mean number of steps necessary to terminate the training. Similarly, we evaluate the mean number of steps when weights are initialized by the Bayesian weights initialization in Algorithm \ref{alg:winit}.

\begin{figure}[H] 
\centering
\includegraphics[scale=0.4]{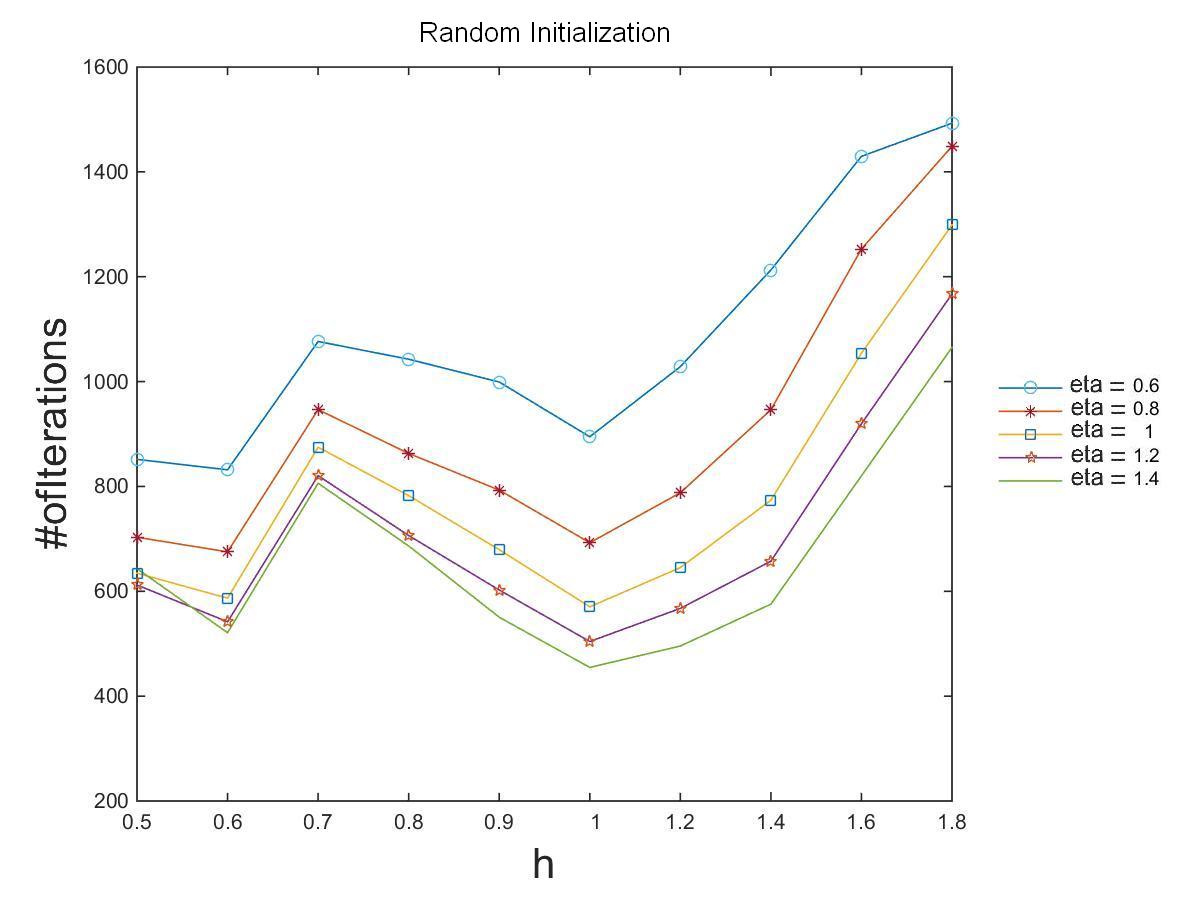}
\caption{Convergence rate of backpropagation algorithm with random weight initialization with $N(2)=70$ applied to recognition of latin printed characters}
\label{fig:N270-hetaB}
\end{figure}

\begin{figure}[H] 
\centering
\includegraphics[scale=0.4]{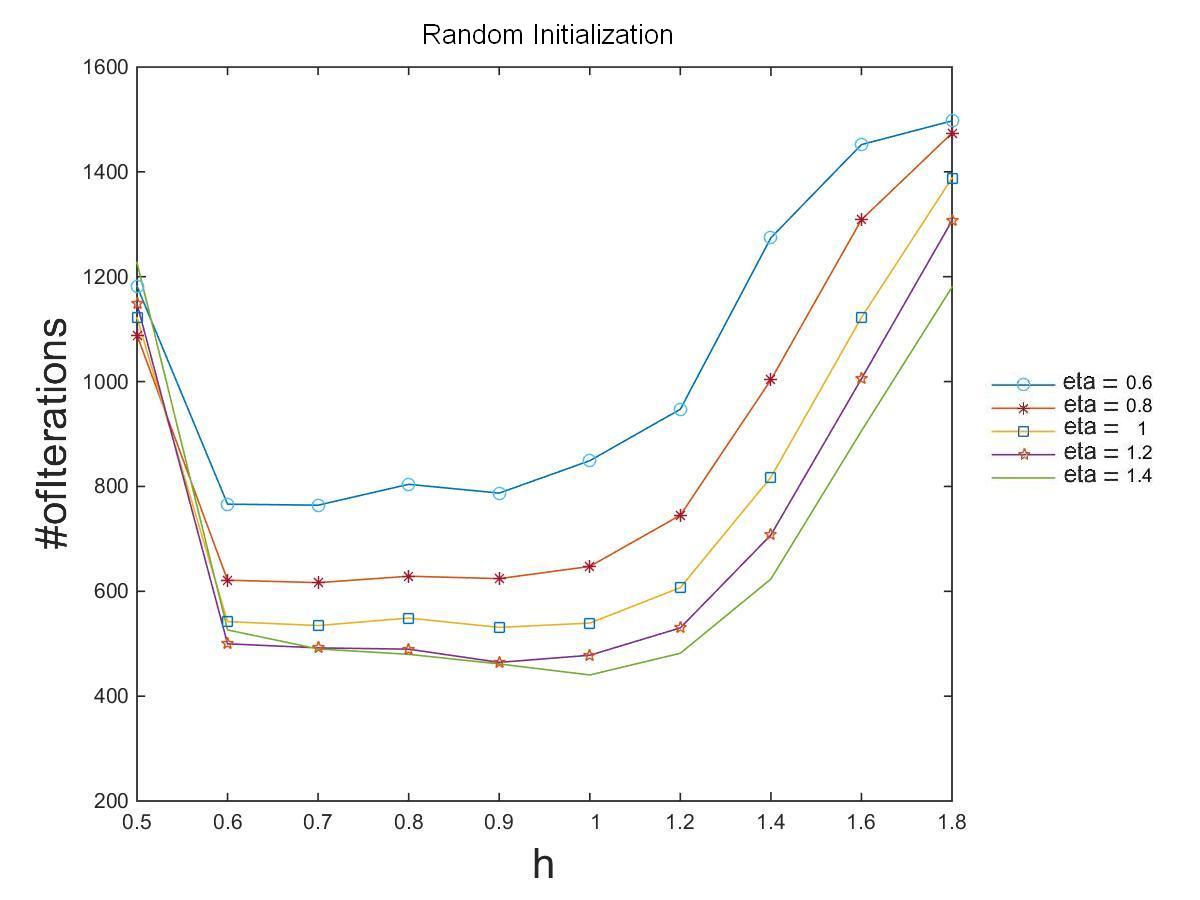}
\caption{Convergence rate of backpropagation algorithm with random weight initialization with $N(2)=80$ applied to recognition of latin printed characters}
\label{fig:N280-hetaB}
\end{figure}

Figures \ref{fig:N270-hetaB} and \ref{fig:N280-hetaB} depict behavior of the mean number of steps that determine convergence of the BP algorithm with RI, for $N(2)=70, 80$, respectively. Each figure reports on the abscissa different values of $h$ and we show the behavior for $\eta=0.6, 0.8, 1, 1.2, 1.4$. 

\begin{figure}[H] 
\centering
\includegraphics[scale=0.4]{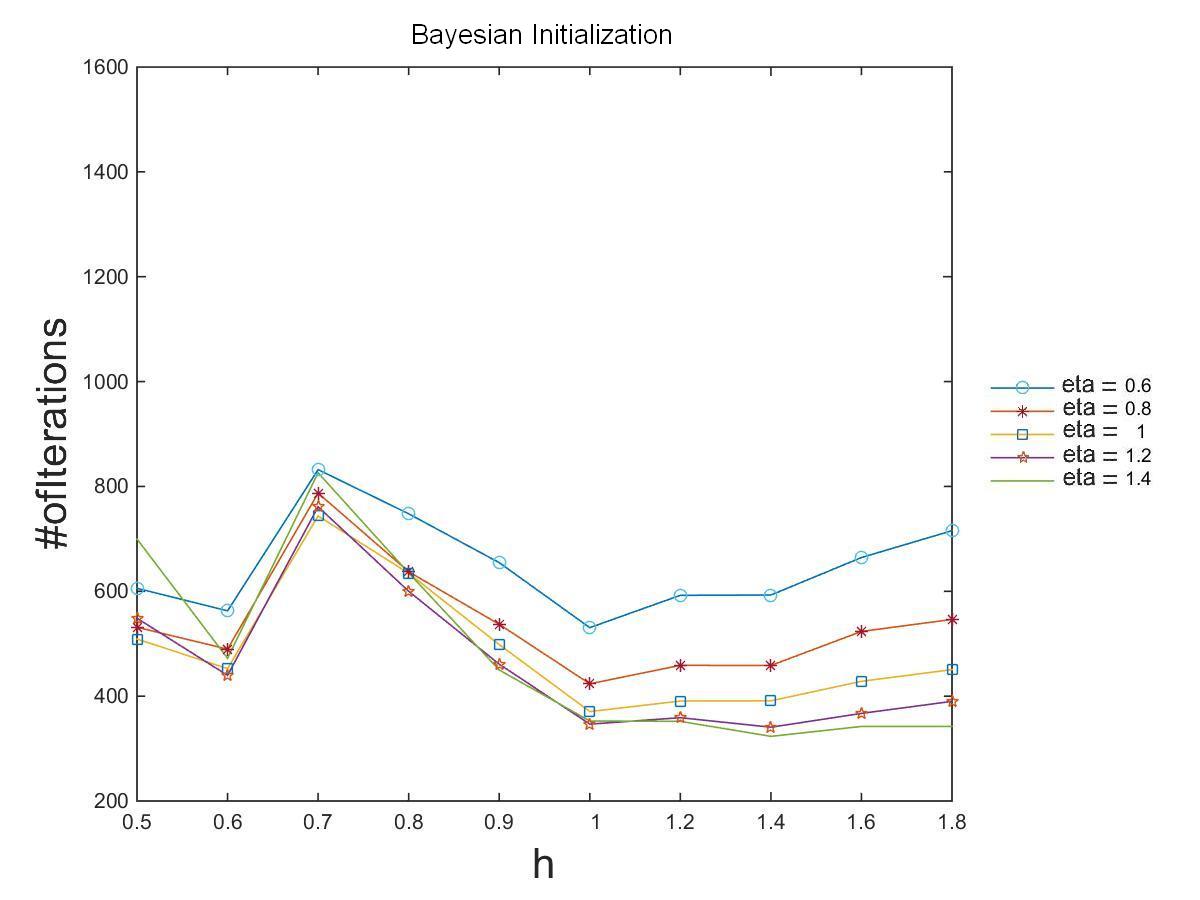}
\caption{Convergence rate of backpropagation algorithm with Bayesian weight initialization with $N(2)=70$ applied to recognition of latin printed characters}
\label{fig:N270-hetaC}
\end{figure}

\begin{figure}[H] 
\centering
\includegraphics[scale=0.4]{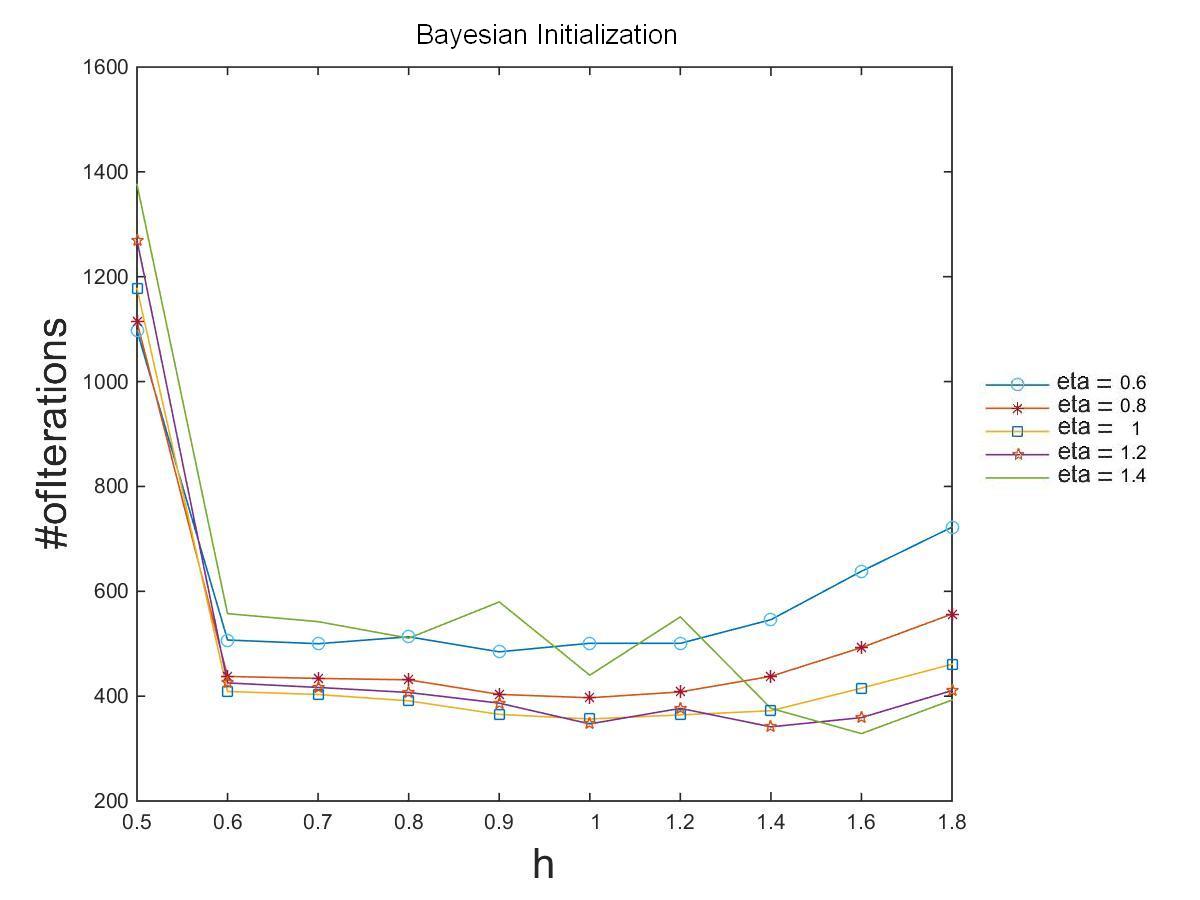}
\caption{Convergence rate of backpropagation algorithm with Bayesian weight initialization with $N(2)=80$ applied to recognition of latin printed characters}
\label{fig:N280-hetaC}
\end{figure}

Figures \ref{fig:N270-hetaC} and \ref{fig:N280-hetaC} show same information for the BP algorithm with BI. 


By figures \ref{fig:N270-hetaB} and \ref{fig:N280-hetaB} (RI), we can observe that for $0.5 \leq h \leq 1$ number of steps, which determine convergence of BP algorithm, generally decreases (with some fluctuation) given any $\eta$. Moreover, increasing values of $\eta$ produce an improvement in the performances. However, such an improvement is less and less noticeable. 

By figures \ref{fig:N270-hetaC} and \ref{fig:N280-hetaC} (BI), we can observe that for $0.5 \leq h \leq 1$ performances of BP algorithm improve, similarly to random initialization. For $h>1$, number of steps, which determine convergence of BP algorithm, increases but slower than the random initialization case. Moreover, increasing values of $\eta$ produce an improvement in the performances, but it is less noticeable than the case of random initialization. 

The improvement in convergence rate due to BI is noticeable at a glance in these figures. In particular, we can see that BI approach is more resistant than RI with respect to high values of $h$, in the sense that number of steps increases slower. In fact, for large values of $h$, weights can range over a large interval. Consequently, RI produces weights scattered on a large interval causing a slower convergence of BP algorithm. On the other hand, BI seems to set initial weights on regions that allow a faster convergence of BP algorithm, despite the size of $h$. This could be very useful in complex problems where small values of $h$ do not allow convergence of BP algorithm and large intervals are necessary.

Moreover, these figures provide some information about optimal values for $h$ and $\eta$ that should be reached around 1 and 1.4, respectively. 

\begin{figure}[hp] 
\centering
\includegraphics[scale=0.4]{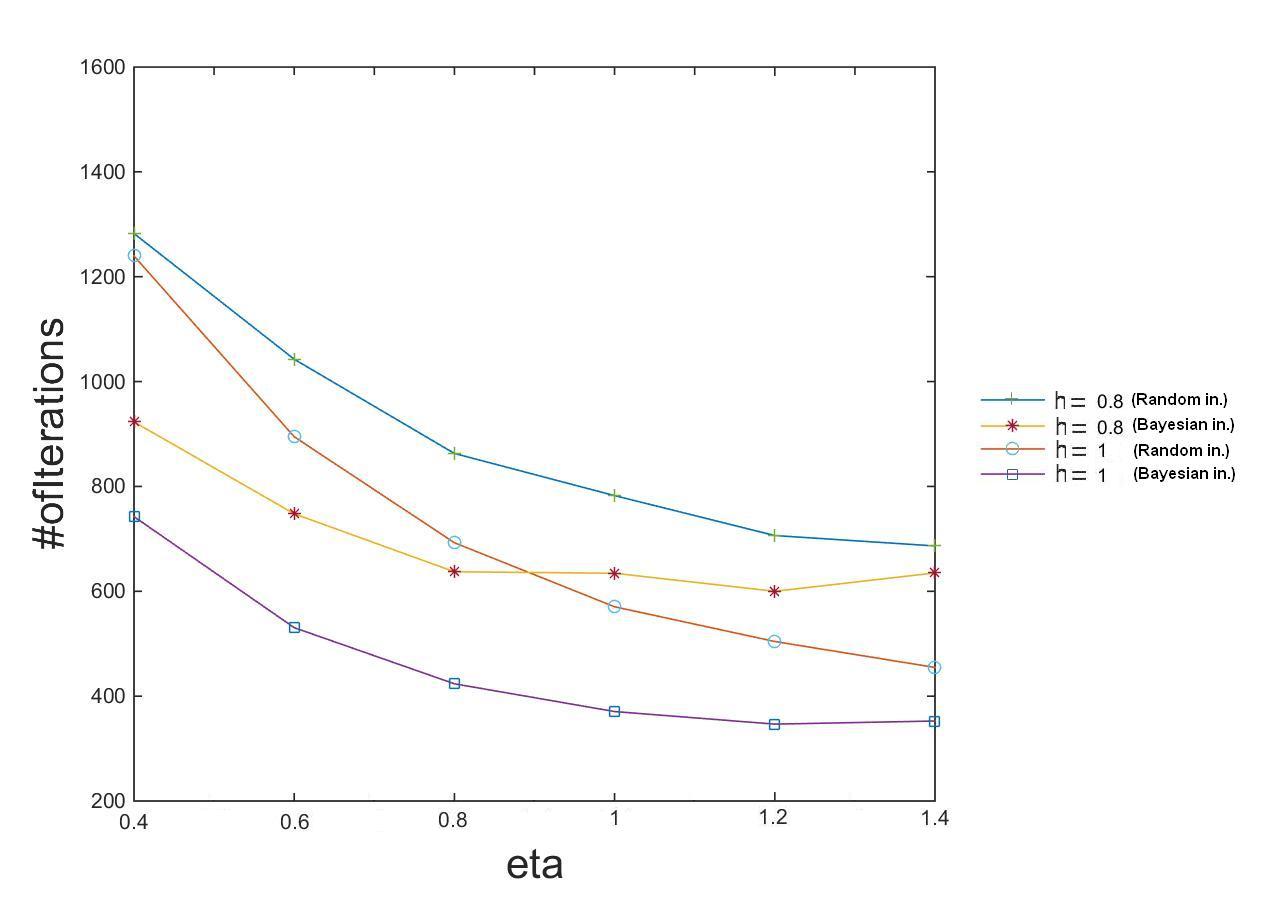}
\caption{Comparison between Bayesian and random weights initialization with $N(2)=70$ and $\eta$ varying on x--axis applied to recognition of latin printed characters}
\label{fig:N270-heta-081-BC}
\end{figure}

\begin{figure}[hp] 
\centering
\includegraphics[scale=0.4]{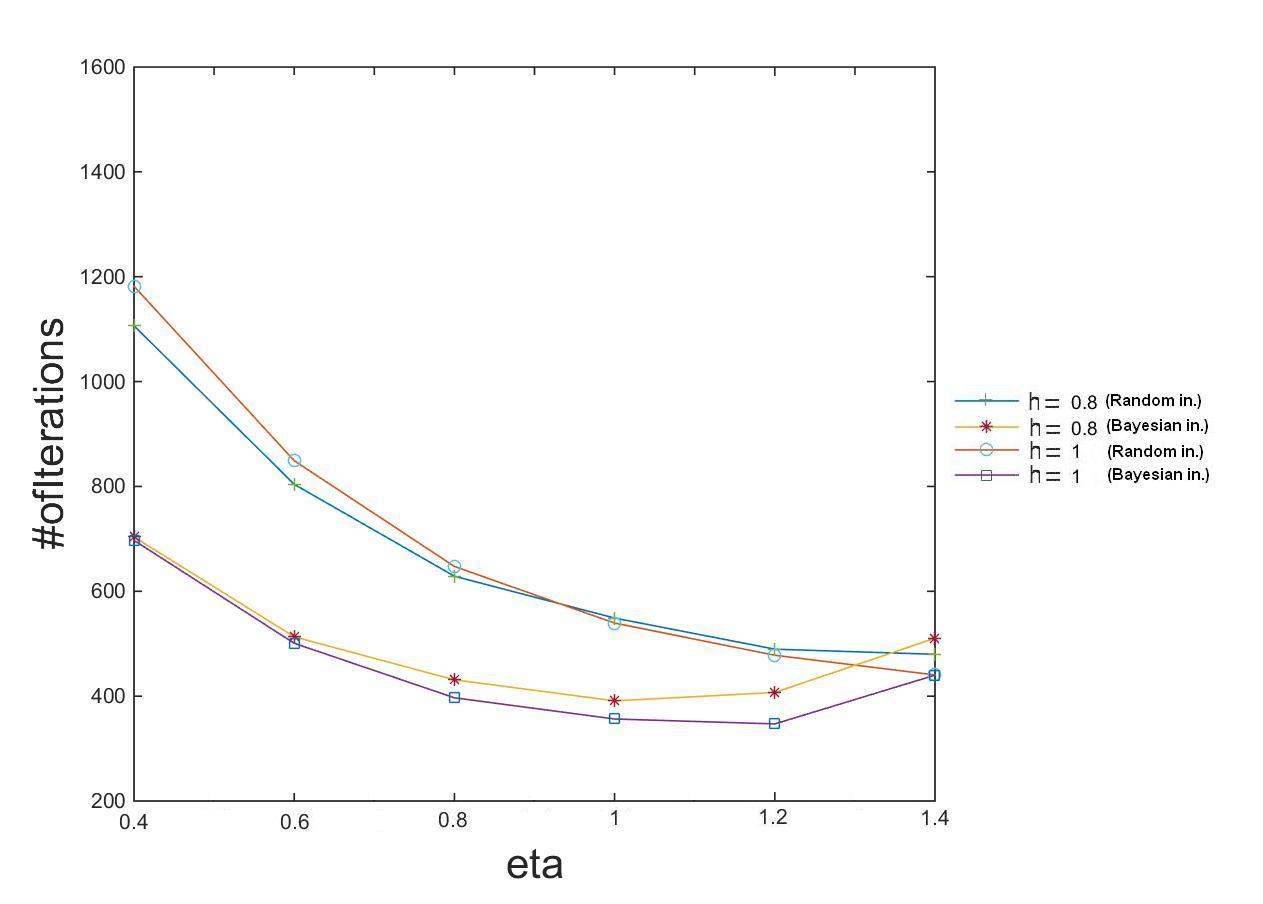}
\caption{Comparison between Bayesian and random weights initialization with $N(2)=80$ and $\eta$ varying on x--axis applied to recognition of latin printed characters}
\label{fig:N280-heta-081-BC}
\end{figure}

\begin{figure}[hp] 
\centering
\includegraphics[scale=0.4]{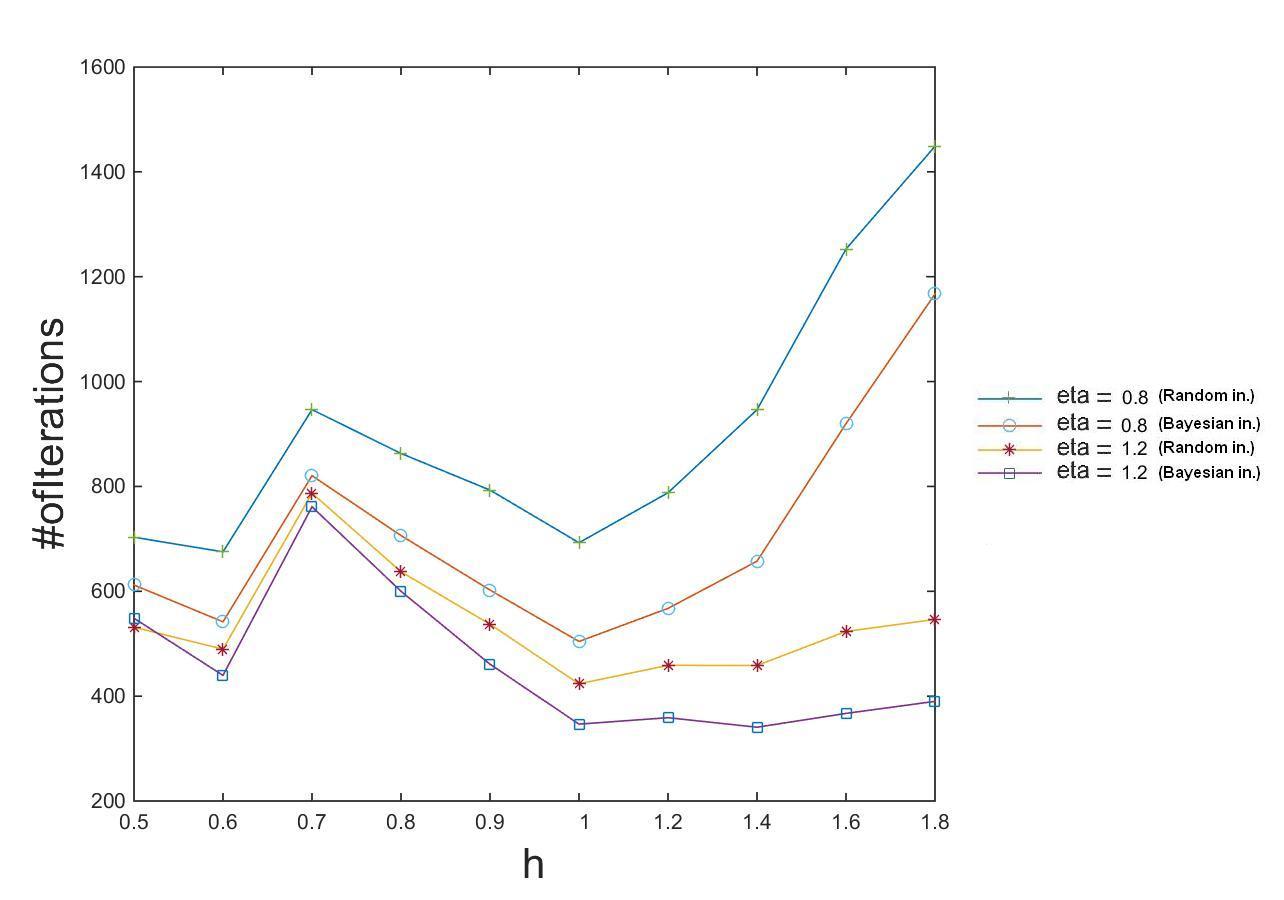}
\caption{Comparison between Bayesian and random weights initialization with $N(2)=70$ and $h$ varying on x--axis applied to recognition of latin printed characters}
\label{fig:N270-heta-0812-BC}
\end{figure}

\begin{figure}[hp] 
\centering
\includegraphics[scale=0.4]{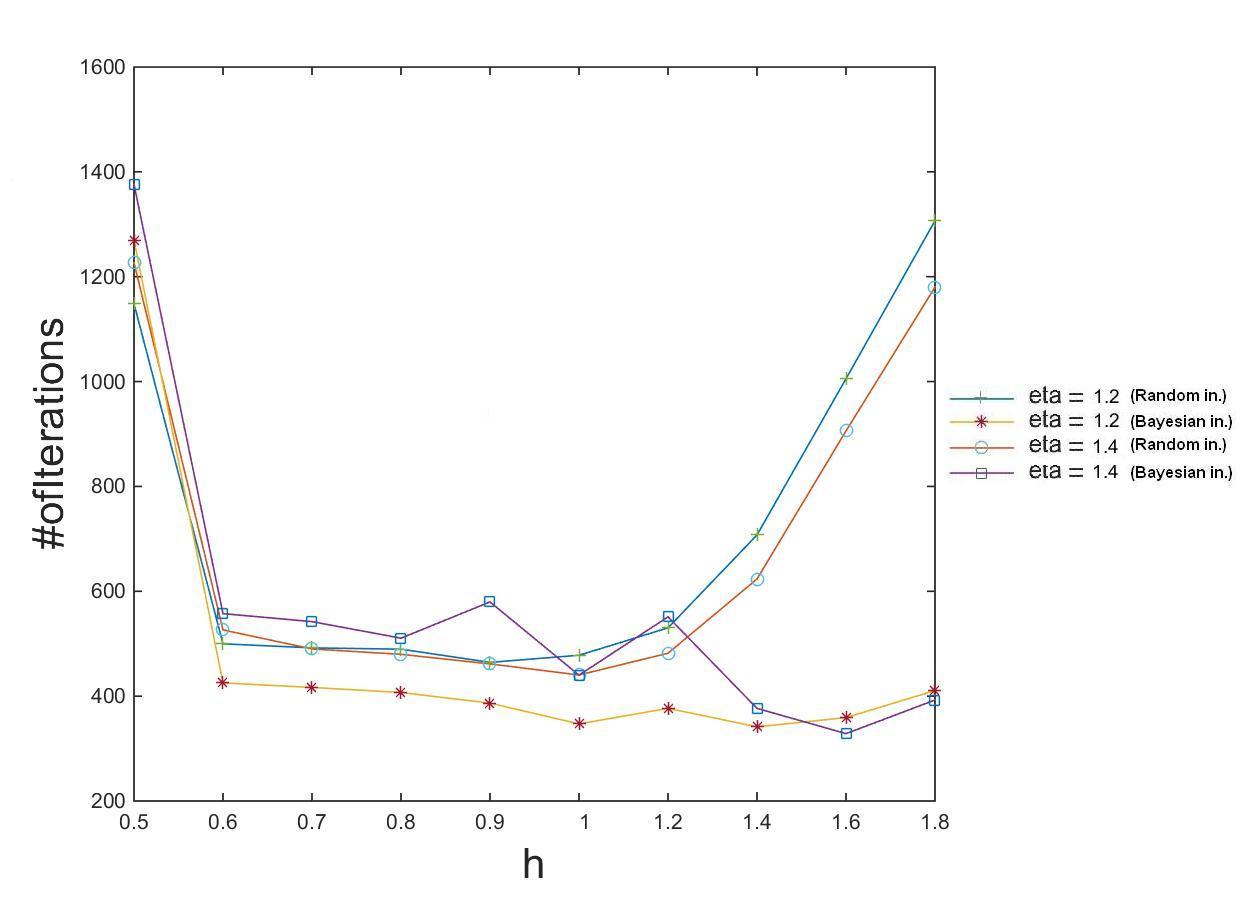}
\caption{Comparison between Bayesian and random weights initialization with $N(2)=80$ and $h$ varying on x--axis applied to recognition of latin printed characters}
\label{fig:N280-heta-1214-BC}
\end{figure}

In Figures \ref{fig:N270-heta-081-BC} and \ref{fig:N280-heta-081-BC} performances of BP algorithm with BI and RI are compared, varying $\eta$ on the x--axis and using two different values for $h$, for $N(2)=70, 80$, respectively. Similarly, figures \ref{fig:N270-heta-0812-BC} and \ref{fig:N280-heta-1214-BC} compare BI and RI, varying $h$ on the x--axis and using two different values for $\eta$, for $N(2)=70, 80$, respectively.

These figures show that generally BI determines an improvement of the convergence rate of the BP algorithm.

In these simulations, the best performance of BP algorithm with RI is obtained with $h=0.9$ and $\eta=1.2$, where the number of steps to terminate the training is 463. The best performance of BP algorithm with BI is obtained with $h=1.6$ and $\eta=1.4$, where the number of steps to terminate the training is 339.

We can observe that for $0.4 \leq \eta \leq 1$, BI improves the convergence rate with respect to RI, given any value of $h$. Furthermore, the improvement of convergence rate is more significant when $h$ increases. For $\eta=1.2$, BI produces improvements only for $h\geq 1.2$, but in this case we can observe that such improvements are significant. For $\eta=1.4$ and $\eta=1.6$, BI produces improvements only for $h=1.4$ and $h=1.6$. Such improvements are very significant both compared to corresponding results obtained by RI and compared to results generally obtained by BI. 

\subsection{Experiments on handwritten digits}\label{sec:mnist}
In this section, we train neural networks in order to recognize handwritten digits in several cases. The benchmark is composed by handwritten digits of the MNIST database. The MNIST database is composed by 60000 handwritten digits usually used as training data and by 10000 handwritten digits usually used as validation data. A handwritten digit is an image with 28 by 28 pixels (gray scale). 

In the following our neural networks have $N_1=28\cdot28=784$ and $N_L=10$. The desired output $\textbf y^{(\textbf x)}$ is the vector $(1,0,0,...,0)$, of length 10, when input $\textbf x$ is the digit 0; it is the vector $(0,1,0,...,0)$ when the input is the digit 1; etc.

For the experiments here presented, we use different neural nets. Specifically, we perform experiments for the following neural nets: $L=3$ and the sigmoidal activation function, $L=3$ and the hyperbolic tangent activation function, $L=5$ and the sigmoidal activation function, $L=5$ and the hyperbolic tangent activation function.

In the above situations, we compare convergence rate of BP algorithm with BI and RI. The convergence rate is evaluated performing 100 different experiments (for each method and situation) and computing the mean value of the steps necessary to achieve the convergence. Moreover, we will also take into account accuracy obtained by these methods testing the trained neural networks on the recognition of 10000 handwritten digits in the MNIST validation test.

\begin{figure}[hp] 
\centering
\includegraphics[scale=0.4]{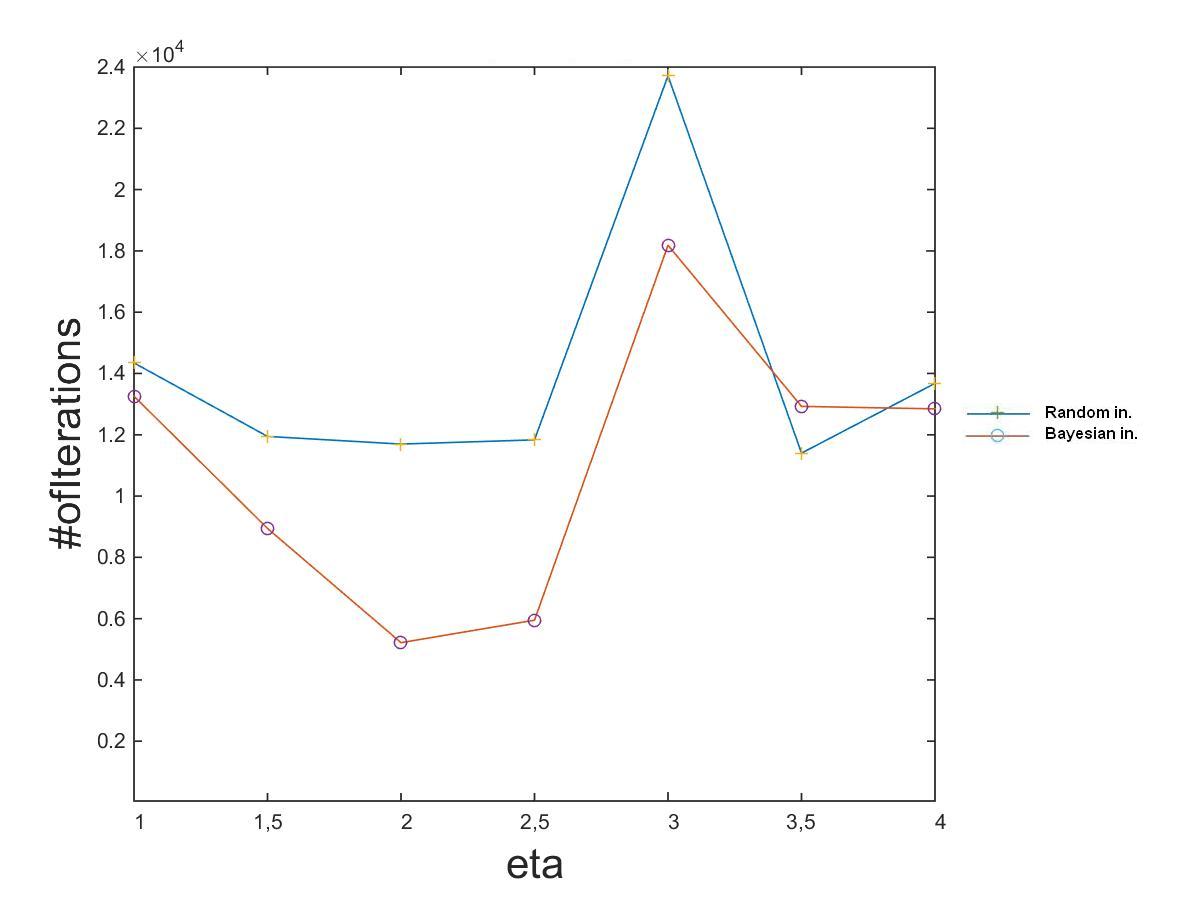}
\caption{Comparison between Bayesian and random weights initialization with $L=3$, $N(2)=70$, $h=1.5$, $\eta$ varying on x--axis, sigmoidal activation function, applied to recognition of handwritten digits of the MNIST database}
\label{fig:MNIST-n20000-h1-sig}
\end{figure}

\begin{figure}[hp] 
\centering
\includegraphics[scale=0.4]{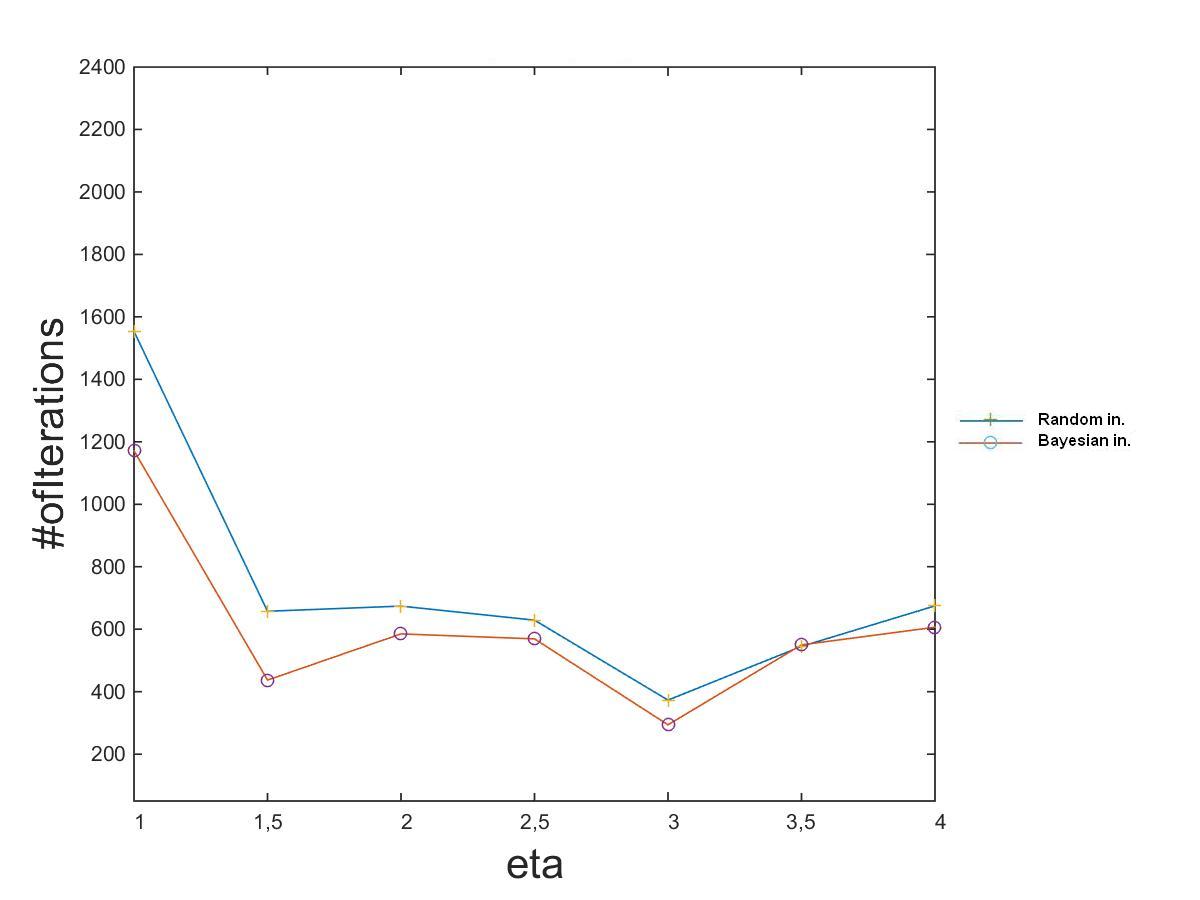}
\caption{Comparison between Bayesian and random weights initialization with $L=3$, $N(2)=70$, $h=1.5$, $\eta$ varying on x--axis, hyperbolic tangent activation function, applied to recognition of handwritten digits of the MNIST database}
\label{fig:MNIST-n20000-h1-htg}
\end{figure}

\begin{figure}[hp] 
\centering
\includegraphics[scale=0.4]{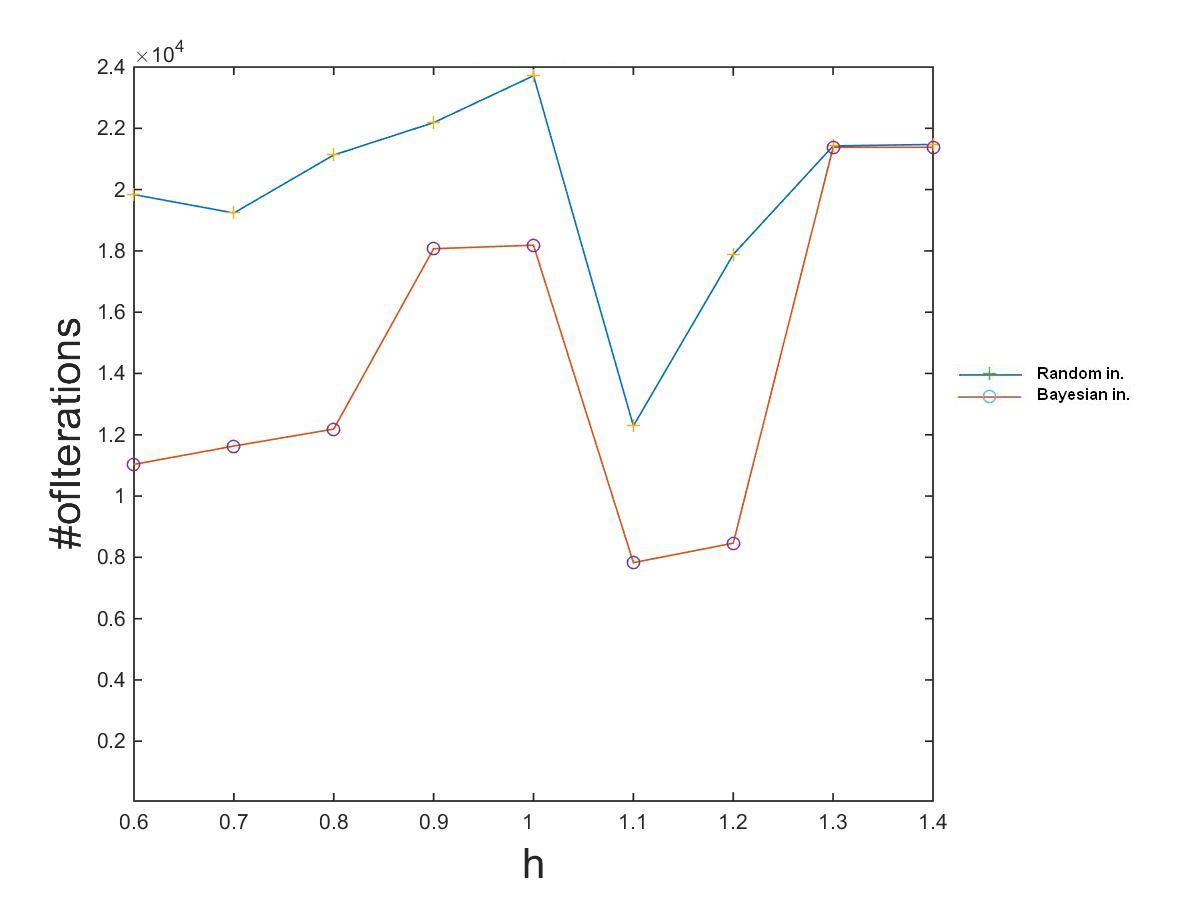}
\caption{Comparison between Bayesian and random weights initialization with $L=3$, $N(2)=70$, $\eta=3.5$, $h$ varying on x--axis, sigmoidal activation function, applied to recognition of handwritten digits of the MNIST database}
\label{fig:MNIST-n20000-eta3-sig}
\end{figure}

\begin{figure}[hp] 
\centering
\includegraphics[scale=0.4]{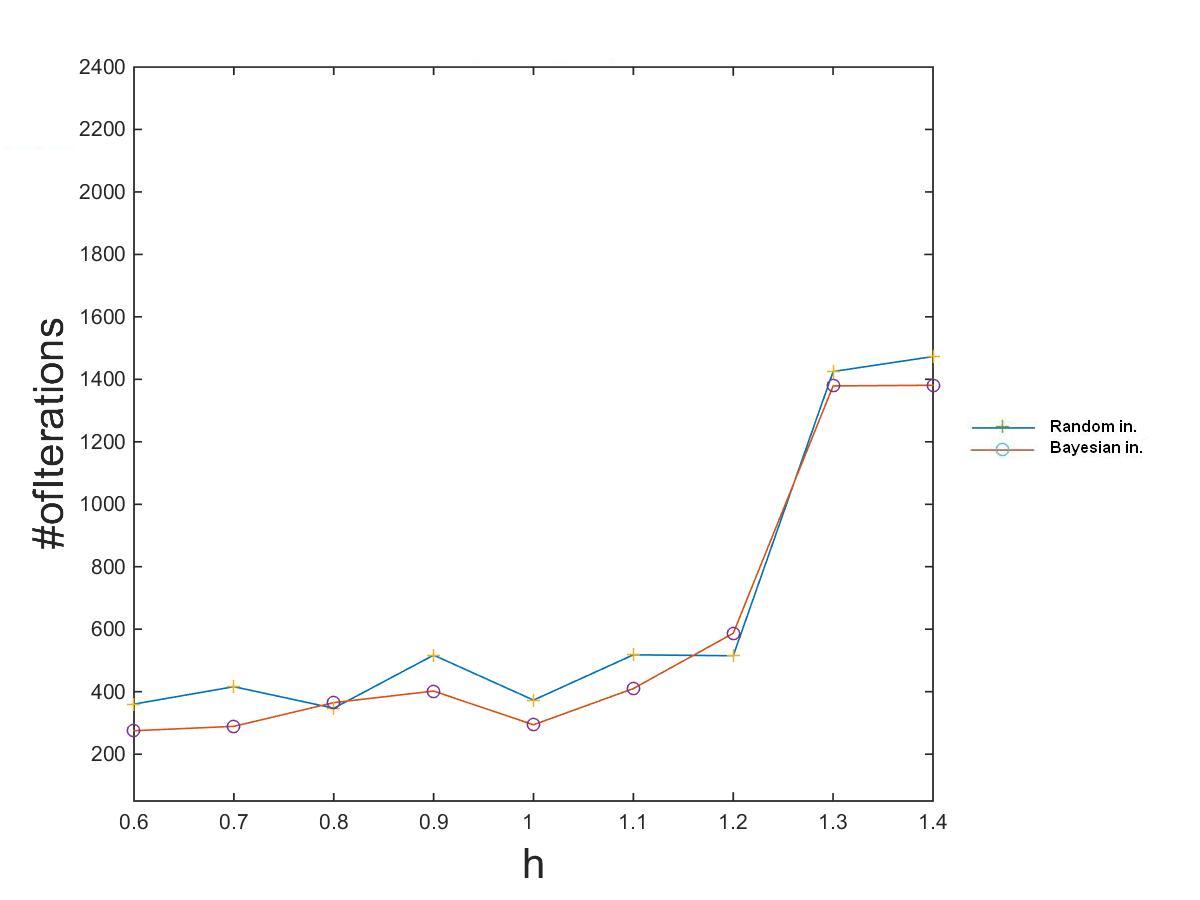}
\caption{Comparison between Bayesian and random weights initialization with $L=3$, $N(2)=70$, $\eta=3.5$, $h$ varying on x--axis, hyperbolic tangent activation function, applied to recognition of handwritten digits of the MNIST database}
\label{fig:MNIST-n20000-eta3-htg}
\end{figure}

In Figures  \ref{fig:MNIST-n20000-h1-sig}, \ref{fig:MNIST-n20000-h1-htg}, \ref{fig:MNIST-n20000-eta3-sig} and \ref{fig:MNIST-n20000-eta3-htg} performances of BP algorithm with BI and RI methods are compared training a neural net with 3 layers, for $N_2=70$, on the first 20000 images contained in the MNIST training set. Specifically, in Figures \ref{fig:MNIST-n20000-h1-sig} and \ref{fig:MNIST-n20000-h1-htg}, we have set $h=1$, varying $\eta$ on the x--axis, and we have used sigmoidal and hyperbolic tangent function, respectively. In Figures \ref{fig:MNIST-n20000-eta3-sig} and \ref{fig:MNIST-n20000-eta3-htg}, we have set $\eta=3.5$, varying $h$ on the x--axis, and we have used sigmoidal and hyperbolic tangent function, respectively. 

\begin{figure}[hp] 
\centering
\includegraphics[scale=0.4]{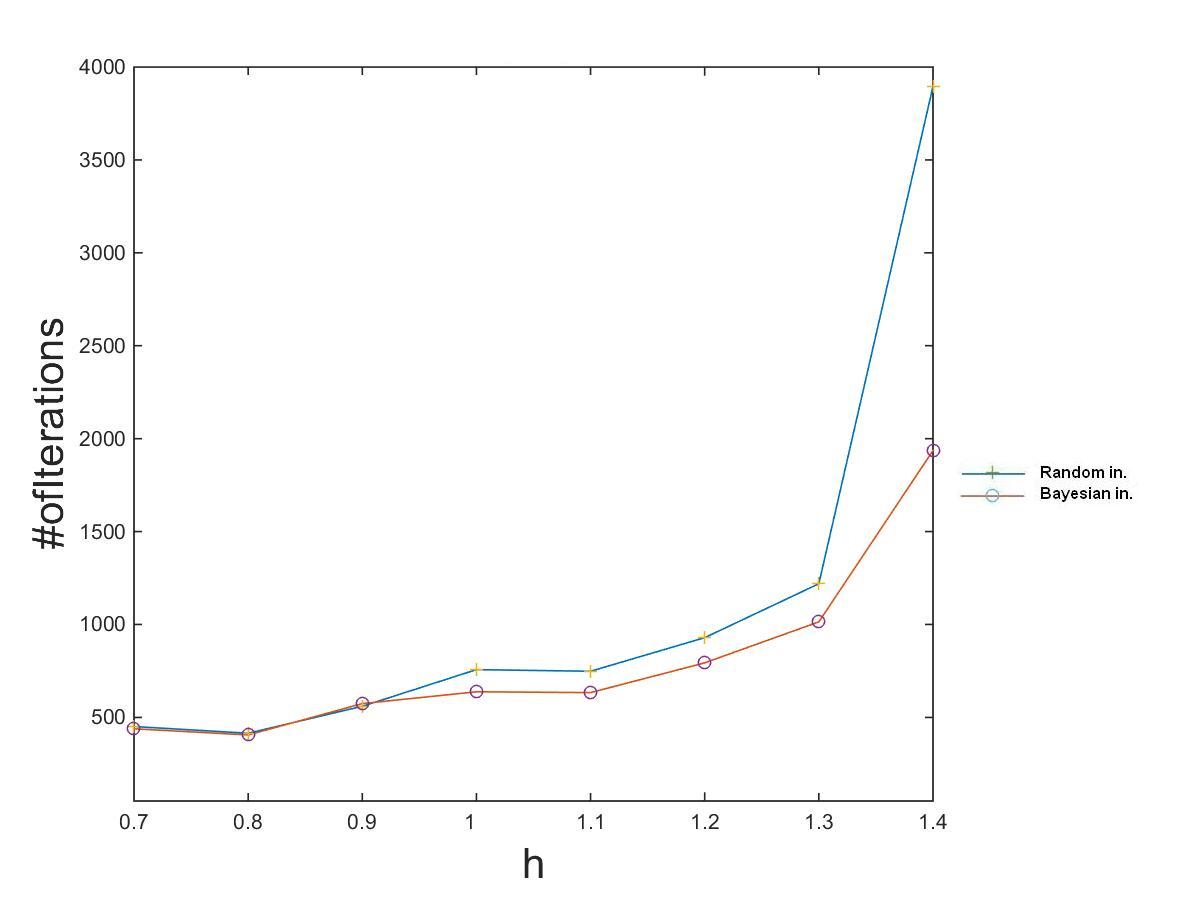}
\caption{Comparison between Bayesian and random weights initialization with $L=5$, $N(2)=50, N(3)=40, N(4)=80$, $\eta=1.4$, $h$ varying on x--axis, hyperbolic tangent activation function, applied to recognition of handwritten digits of the MNIST database}
\label{fig:MNIST-n20000-L5-eta14}
\end{figure}

\begin{figure}[hp] 
\centering
\includegraphics[scale=0.4]{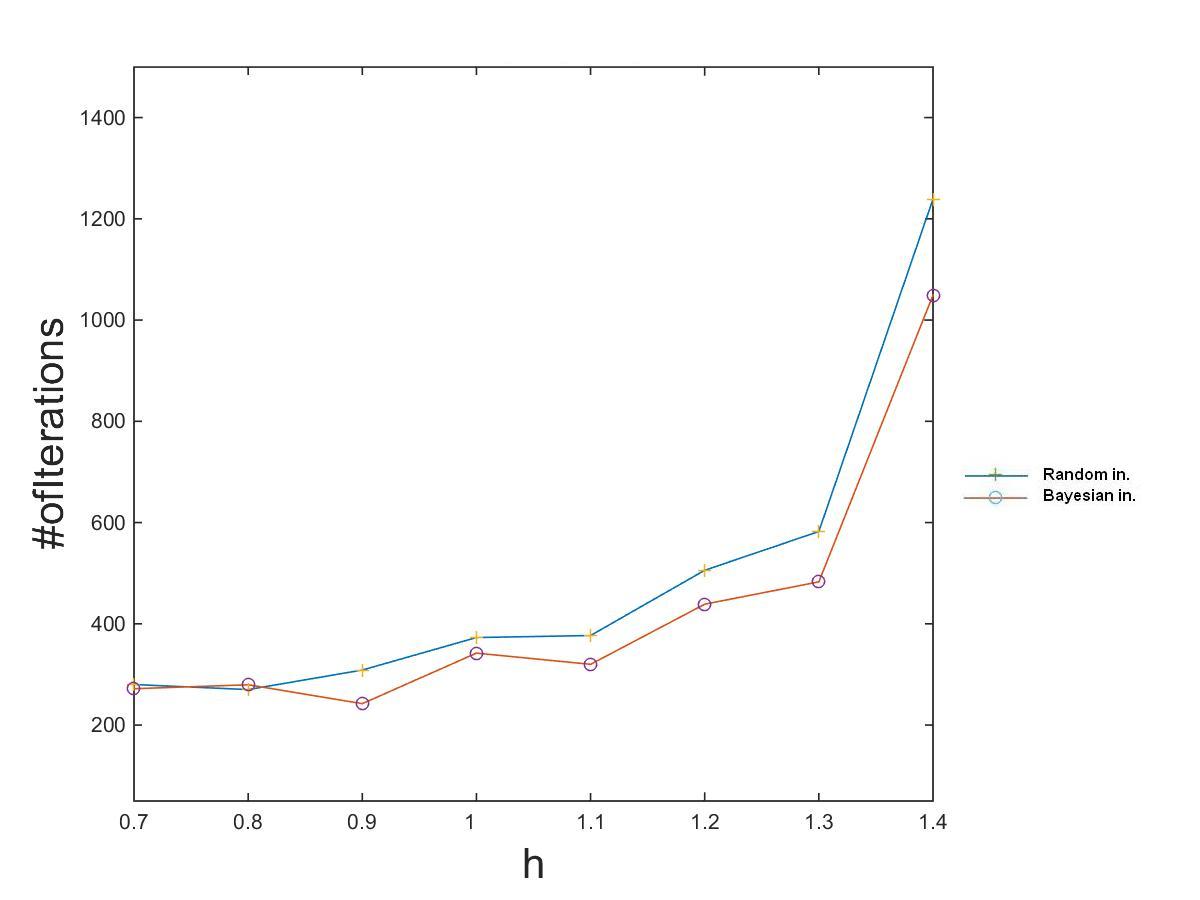}
\caption{Comparison between Bayesian and random weights initialization with $L=5$, $N(2)=50, N(3)=40, N(4)=80$, $\eta=2.5$, $h$ varying on x--axis, hyperbolic tangent activation function, applied to recognition of handwritten digits of the MNIST database}
\label{fig:MNIST-n20000-L5-eta25}
\end{figure}

\begin{figure}[H] 
\centering
\includegraphics[scale=0.4]{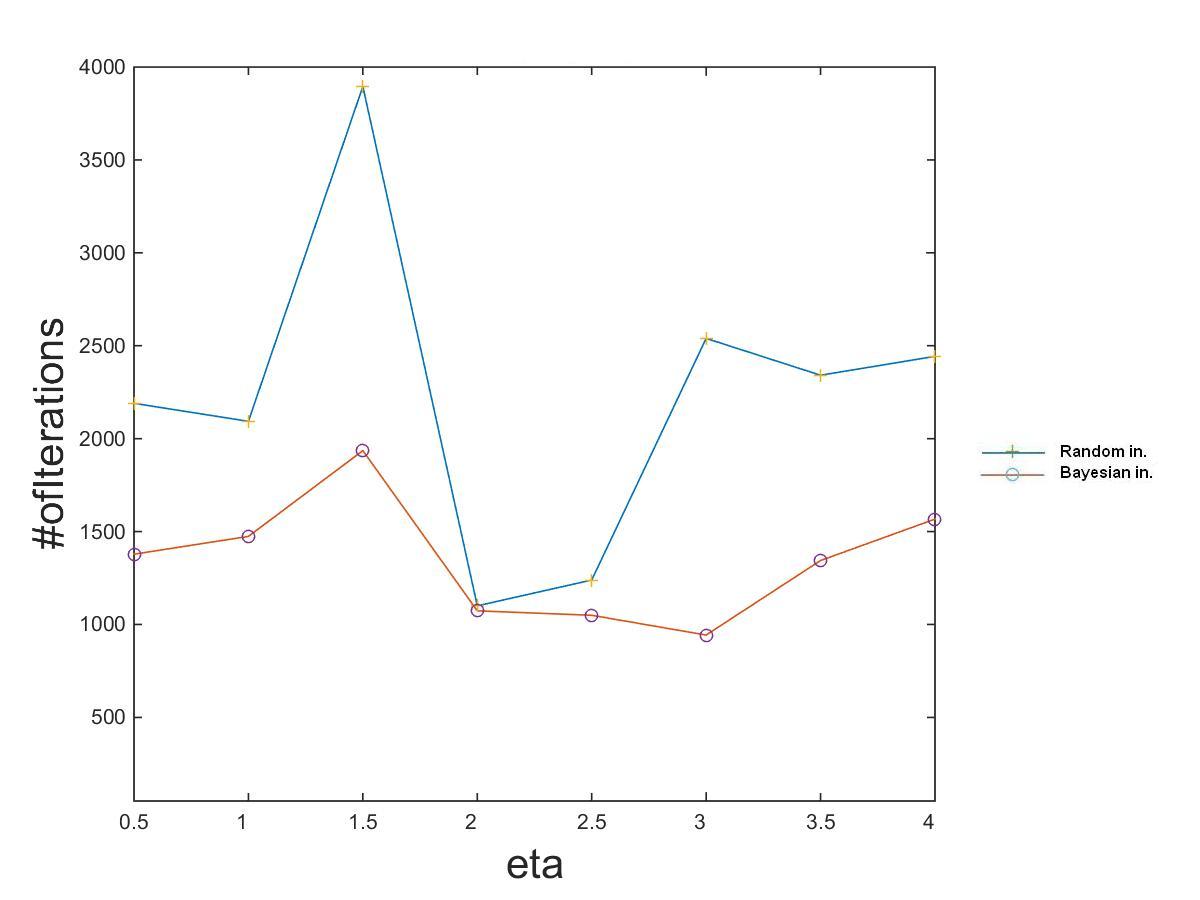}
\caption{Comparison between Bayesian and random weights initialization with $L=5$, $N(2)=50, N(3)=40, N(4)=80$, $h=1.4$, $\eta$ varying on x--axis, hyperbolic tangent activation function, applied to recognition of handwritten digits of the MNIST database}
\label{fig:MNIST-n20000-L5-h14}
\end{figure}

Figures \ref{fig:MNIST-n20000-L5-eta14}, \ref{fig:MNIST-n20000-L5-eta25} and \ref{fig:MNIST-n20000-L5-h14} show behavior of BP algorithm with BI and RI methods for a neural network with 5 layers. We have set $N_2=50$, $N_3=40$, $N_4=80$ (note that this parameters have not been optimized, thus different deep neural nets could obtain better performances). In Figures \ref{fig:MNIST-n20000-L5-eta14}, \ref{fig:MNIST-n20000-L5-eta25}, we vary $h$ on the x--axis for $\eta=1.4$ and $\eta=2.5$, respectively. In Figure \ref{fig:MNIST-n20000-L5-h14}, we vary $\eta$ on the x--axis for $h=1.4$. For all the above situations we have used the hyperbolic tangent as activation function.

These experiments confirm performances observed in the previous sections. Indeed, BI generally determines an improvement of the convergence rate of the BP algorithm with respect to RI.

In Figures \ref{fig:MNIST-n20000-h1-sig} and \ref{fig:MNIST-n20000-h1-htg}, BI has a worst performance than RI only for $\eta=3.5$ and we can observe that we have significant improvement of convergence rate for low values of $\eta$. Thus, in Figures \ref{fig:MNIST-n20000-eta3-sig} and \ref{fig:MNIST-n20000-eta3-htg} we have tested our method in situations where it seems to have poor performances (i.e, for high values of $\eta$). Specifically we used $\eta=3.5$, varying $h$ on x--axis from 0.6 to 1.4. In these simulations, the results are good: for $h\leq 1.2$ BI determines a faster convergence than RI. Moreover, let us observe that best performances are generally obtained when $\eta\leq3$ and $h\leq1.2$ for both BI and RI. Thus, the use of high values of $\eta$ and $h$ is not suitable in this context.

Figures \ref{fig:MNIST-n20000-L5-eta14}, \ref{fig:MNIST-n20000-L5-eta25} and \ref{fig:MNIST-n20000-L5-h14} show that BI method generally improves convergence rate of BP algorithm also when deep neural networks are used. In this cases, we see that when $h$ increases, the distance between number of steps to achieve convergence with BI and RI is more marked in favor of BI method.  

Finally, in Tables \ref{table:accuracy20000} and \ref{table:accuracy60000} we have analyzed classification accuracy of neural networks trained using the BI against RI.

\begin{table}[ht]\small  
\centering
\tabcolsep=0.15cm
\scalebox{0.82}{
\begin{tabular}{|c||c|c|c|c|c|c|c|c|}
\hline 
 & \multicolumn{4}{|c|}{$L=5$, $\eta=1.5$} & \multicolumn{4}{c|}{$L=3$, $\eta=3$}\\
 \hline
 & \multicolumn{2}{|c|}{Random in.} & \multicolumn{2}{|c|}{Bayes in.} & \multicolumn{2}{|c|}{Random in.} & \multicolumn{2}{c|}{Bayes in.}\\
 \hline
h  & Steps & Perc. rec. & Steps & Perc. rec. & Steps & Perc. rec. & Steps & Perc. rec. \cr \hline
0.7 & 451 & 85 & 439 & 87 & 416 & 91 & 289 & 92 \cr \hline
0.8 & 415 & 81 & 405 & 85 & 347 & 92 & 365 & 92 \cr \hline 
0.9 & 560 & 81 & 574 & 82 & 517 & 90 & 402 & 92 \cr \hline
1 &   757 & 78 & 638 & 79 & 373 & 88 & 294 & 91 \cr \hline
1.1 & 748 & 86 & 633 & 86 & 518 & 92 & 410 & 92 \cr \hline
1.2 & 929 & 80 & 793 & 80 & 515 & 90 & 587 & 91 \cr \hline
1.3 & 1219 & 82 & 1014 & 81 & 1425& 85 & 1379& 89 \cr \hline
1.4 & 3896& 77 & 1936 & 76 & 1473& 81 & 1381& 81 \cr \hline
\end{tabular}
}
\caption{Percentage of recognized digits in the MNIST validation set. Neural networks trained on first 20000 digits of the MNIST training set.}
\label{table:accuracy20000}
\end{table}

\begin{table} [ht]\small
\centering
\tabcolsep=0.15cm
\scalebox{0.82}{
\begin{tabular}{|c||c|c|c|c|}
\hline 
 & \multicolumn{2}{|c|}{$L=5$, $h=0.8$} & \multicolumn{2}{c|}{$L=3$, $h=1$}\\
 \hline
$\eta$ & Random in. & Bayes in. & Random in. & Bayes in. \cr \hline
0.5 & 92 & 92 & 90 & 91 \cr \hline
1 & 92 & 95 & 91 & 94 \cr \hline 
1.5 & 95 & 96 & 94 & 93 \cr \hline
2 & 93 & 95 & 93 & 93 \cr \hline
2.5 & 92 & 96 & 95 & 94 \cr \hline
3 & 90 & 88 & 95 & 96 \cr \hline
\end{tabular}
}
\caption{Percentage of recognized digits in the MNIST validation set. Neural networks trained for 300 steps on the MNIST training set.}
\label{table:accuracy60000}
\end{table}

In Table \ref{table:accuracy20000}, we have tested neural networks in the recognition of the 10000 digits of the MNIST validation set, when the training on the first 20000 digits of the MNIST training set is terminated. In Table \ref{table:accuracy60000}, we have tested neural networks in the recognition of the 10000 digits of the MNIST validation set, after 300 steps of training on the 60000 digits of the MNIST training set. We have chosen to perform these simulations in order to highlight differences in terms of accuracy between BI and RI methods. In the case of the MNIST database, if training is accomplished over all the training dataset, then BP algorithm for multilayer neural networks yields a very high accuracy (more than $99\%$, see, e.g., \cite{Cir}) and consequently differences in terms of accuracy are hard to see.

We can observe that percentage of recognized digits is generally greater when BI is used. This result could be expected for simulations reported in Table \ref{table:accuracy60000}, since after the same number of steps the neural network with BI recognizes a greater number of digits of the training step than neural network with RI (since, neural net with BI converges faster than neural net with RI). Moreover, these results are also confirmed in Table \ref{table:accuracy20000} where both neural net with BI and RI have terminated the training.

\subsection{Comparison with other initialization methods}\label{sec:comparison}
In this section, we compare performances of BI method with other ones. We use results provided in \cite{Red}, where several methods have been tested and compared on different benchmarks from the UCI repository of machine learning databases. Specifically, we perform tests on the following problems: Balance Scale (BAL), Cylinders Bands (BAN), Liver Disorders (LIV), Glass Identification (GLA), Heart Disease (HEA), Imagen Segmentation (IMA). The methods tested in \cite{Red} have been developed by Drago and Ridella \cite{Dra} (Method A), Kim and Ra \cite{Kim} (Method B), Palubinskas \cite{Pal} (Method C), Shimodaira \cite{Shi} (method D), Yoon et al. \cite{Yoon} (Method E). Note that in \cite{Red} these methods are labeled in a different way.

In Table \ref{table:comparison}, we report the mean number of steps to achieve convergence with the BP algorithm (30 different trials are performed). Results of Methods A, B, C, D, E and RI are reported from \cite{Red}. The BI method is tested with $h=0.05$ (since in \cite{Red} weights are sampled in the interval $[-0.05,0.05]$) and $\eta=2$. Tables \ref{table:comparison-noconv} and \ref{table:comparison-perc} also provide the number of trials where algorithms do not achieve convergence and the mean percentage of correct recognitions after training, respectively.

\begin{table}[hp]\small

\centering
\tabcolsep=0.15cm
\scalebox{0.82}{
\begin{tabular}{|c||c|c|c|c|c|c|}
\hline 
   \backslashbox{Method}{Problem}  & \textbf{BAL} & \textbf{BAN} & \textbf{LIV} & \textbf{GLA} & \textbf{HEA} & \textbf{IMA}    \cr \hline \hline
 \textbf{Meth. RI} & 120 & 800 & 1300 & 111 & 220 & 710  \cr \hline
 \textbf{Meth. A} & 130 & 600 & 1600 & 230 & 200 & 1090 \cr \hline
 \textbf{Meth. B} & 80 & 720 & 1300 & 150 & 320 & 1010 \cr \hline
 \textbf{Meth. C} & 120 & 700 & 2800 & 160 & 430 & 950  \cr \hline
 \textbf{Meth. D} & 80 & 470 & 500 & 91 & 290 & 970 \cr \hline
 \textbf{Meth. E} & 270 & 800 & 2100 & 300 & 500 & 1040  \cr \hline
 \textbf{Meth. BI} & 89 & 523 & 925 & 84 & 161 & 459 \cr \hline
 
\end{tabular}
}
\caption{Mean number of steps of backpropagation algorithm to converge with different initialization methods applied to different problems.}

\label{table:comparison}
\end{table}

\begin{table}[hp]\small

\centering
\tabcolsep=0.15cm
\scalebox{0.82}{
\begin{tabular}{|c||c|c|c|c|c|c|}
\hline 
   \backslashbox{Method}{Problem}  & \textbf{BAL} & \textbf{BAN} & \textbf{LIV} & \textbf{GLA} & \textbf{HEA} & \textbf{IMA}    \cr \hline \hline
 \textbf{Meth. RI} & 1 & 11 & 3 & 3 & 4 & 5  \cr \hline
 \textbf{Meth. A} & 1 & 5 & 4 & 3 & 4 & 5 \cr \hline
 \textbf{Meth. B} & 0 & 8 & 1 & 2 & 3 & 4 \cr \hline
 \textbf{Meth. C} & 0 & 8 & 3 & 2 & 4 & 5  \cr \hline
 \textbf{Meth. D} & 0 & 4 & 0 & 2 & 2 & 7 \cr \hline
 \textbf{Meth. E} & 0 & 5 & 4 & 7 & 3 & 11  \cr \hline
 \textbf{Meth. BI} & 0 & 9 & 4 & 3 & 3 & 2 \cr \hline
 
\end{tabular}
}
\caption{Number of non--convergent trials for backpropagation algorithm with different initialization methods applied to different problems.}
\label{table:comparison-noconv}
\end{table}

\begin{table}[hp]\small

\centering
\tabcolsep=0.15cm
\scalebox{0.82}{
\begin{tabular}{|c||c|c|c|c|c|c|}
\hline 
   \backslashbox{Method}{Problem}  & \textbf{BAL} & \textbf{BAN} & \textbf{LIV} & \textbf{GLA} & \textbf{HEA} & \textbf{IMA}    \cr \hline \hline
 \textbf{Meth. RI} & 91.8 & 66.8 & 59.4 & 90.4 & 81.2 & 72  \cr \hline
 \textbf{Meth. A} & 90.6 & 67.7 & 60.9 & 88.2 & 80.8 & 70 \cr \hline
 \textbf{Meth. B} & 91 & 66.9 & 60 & 88.9 & 81.7 & 70 \cr \hline
 \textbf{Meth. C} & 91.1 & 68.3 & 60.8 & 90.7 & 80.6 & 74.7  \cr \hline
 \textbf{Meth. D} & 91.7 & 68.5 & 63.1 & 91.9 & 81.7 & 76 \cr \hline
 \textbf{Meth. E} & 91.4 & 65.3 & 61.3 & 85.7 & 80.9 & 59  \cr \hline
 \textbf{Meth. BI} & 91.3 & 69.1 & 62.6 & 89.2 & 81.4 & 71.8 \cr \hline
 
\end{tabular}
}
\caption{Percentage of correct recognition after training by backpropagation algorithm converge with different initialization methods applied to different problems.}
\label{table:comparison-perc}
\end{table}

In terms of convergence rate, we can see that methods B and D have better performances than BI method in Balance Scale problem, whereas only method D converges faster than BI in Cylinder Bands and Liver Disorders problems. In the remaining problems, BI method provides the best performances. On the other hand, in these trials we can not observe significant improvements of the BI method with respect to RI and other methods about mean percentage of correct recognitions and number of trials not achieving convergence. We can observe that BI method generally improves RI, but the results of these tests can not be considered significant, since similar results are reached.

We can observe that, as stated in \cite{Red}, Method D needs determining several parameters by a trial and error procedure. Indeed, here we only reported the best performances of Method D obtained in \cite{Red}, where the method is tested with several different values of the parameters. On the contrary, BI method does not need tuning extra parameters.

\section{Conclusion and future work} \label{sec:conc}
In this paper, the problem of convergence rate of the backpropagation algorithm for training neural networks has been treated. A novel method for the initialization of weights in the backpropagation algorithm has been proposed. The method is mainly based on an innovative use of the Kalman filter with an original metrological approach. A simulation study has been carried on to show the benefits of the proposed method with respect to random weights initialization, applying the neural net in the field of the character recognition. Some comparisons with other initialization methods have been performed. The obtained results are encouraging, and we expect that the new features we introduced are actually relevant in a variety of application contexts of neural nets. In particular, the Bayesian weights initialization could be very useful to solve complex problems where weights need large values of $h$ to ensure convergence of BP algorithm.
Looking at perspective advancements, the following issues could be addressed in future works:
\begin{itemize}
\item values of entries of covariance matrix $R_t(k)$ should be further optimized by means of a deeper study on correlations among weights of neural networks;
\item theoretical analysis of the convergence of the BP algorithm with BI, evaluating and comparing the initial expected error of the neural network whose weights are initialized with the Bayesian approach against the expected error due to random initialization;
\item application of the BI method to complex problems needing large values of $h$;
\item recently, the greedy layer--wise unsupervised pre--training has been introduced in order to achieve fast convergence for backpropagation neural networks \cite{Hin}, \cite{Ben}, \cite{Erh}; it could be interesting to compare this method with BI. Moreover, BI could be exploited in order to improve the pre--training of this method. In fact, the greedy layer--wise unsupervised pre--training involves several operations to initialize the weights in the final/overall deep network. Moreover, the random initialization of weights of neural nets is still the more widespread method. Thus, the study of simple methods that improves random initialization, like the Bayesian approach proposed here, is still an active research field.
\end{itemize}

\section{Acknowledgments}
This work has been developed in the framework of an agreement between IRIFOR/UICI (Institute for Research, Education and Rehabilitation/Italian Union for the Blind and Partially Sighted) and Turin University.

Special thanks go to Dott. Tiziana Armano and Prof. Anna Capietto for their support to this work.

We would like to thanks the anonymous referees whose suggestions have improved the paper.

\end{document}